\documentclass{article}

\PassOptionsToPackage{numbers, compress}{natbib}
\usepackage[preprint]{neurips_2020}

\usepackage[utf8]{inputenc} % allow utf-8 input
\usepackage[T1]{fontenc}    % use 8-bit T1 fonts
\usepackage{hyperref}       % hyperlinks
\usepackage{url}            % simple URL typesetting
\usepackage{booktabs}       % professional-quality tables
\usepackage{amsfonts}       % blackboard math symbols
\usepackage{nicefrac}       % compact symbols for 1/2, etc.
\usepackage{microtype}      % microtypography
\usepackage[utf8]{inputenc}
\usepackage[ruled, vlined]{algorithm2e}
\usepackage[mathscr]{euscript}
\usepackage{amsmath}
\usepackage{amsthm}
\usepackage{amssymb}
\usepackage{graphicx}
\usepackage{subfig}
\usepackage{float}
\usepackage{caption}
\usepackage{verbatim}
\usepackage{cleveref}
\usepackage{wrapfig}
\usepackage{adjustbox}
\usepackage{color}

\usepackage{amsthm}

\theoremstyle{definition}
\newtheorem{definition}{Definition}[section]

\title{Supervised Visualization for Data Exploration}
\author{Jake S. Rhodes\\
        Department of Mathematics \& Statistics\\
        Utah State University\\
        \texttt{jakerhodes@aggiemail.usu.edu}\\
        \And
        Adele Cutler\\
        Department of Mathematics \& Statistics\\
        Utah State University\\
        \texttt{adele.cutler@usu.edu}
        \And
        Guy Wolf\\
        Department of Mathematics and Statistics\\
        Universit\'{e} de Montr\'{e}al\\
        \texttt{guy.wolf@umontreal.ca}
        \And
        Kevin R. Moon\\
        Department of Mathematics \& Statistics\\
        Utah State University\\
        \texttt{kevin.moon@usu.edu}
        }

\begin{document}

\maketitle
\begin{abstract}
    Dimensionality reduction is often used as an initial step in data exploration, either as preprocessing for classification or regression or for visualization.  Most dimensionality reduction techniques to date are unsupervised; they do not take class labels into account (e.g., PCA, MDS, t-SNE, Isomap). Such methods require large amounts of data and are often sensitive to noise that may obfuscate important patterns in the data. Various attempts at supervised dimensionality reduction methods that take into account auxiliary annotations (e.g., class labels) have been successfully implemented with goals of increased classification accuracy or improved data visualization. Many of these supervised techniques incorporate labels in the loss function in the form of similarity or dissimilarity matrices, thereby creating over-emphasized separation between class clusters, which does not realistically represent the local and global relationships in the data. In addition, these approaches are often sensitive to parameter tuning, which may be difficult to configure without an explicit quantitative notion of visual superiority. In this paper, we describe a novel supervised visualization technique based on random forest proximities and diffusion-based dimensionality reduction. We show, both qualitatively and quantitatively, the advantages of our approach in retaining local and global structures in data, while emphasizing important variables in the low-dimensional embedding. Importantly, our approach is robust to noise and parameter tuning, thus making it simple to use while producing reliable visualizations for data exploration.
\end{abstract}

\section{Introduction}\label{section:intro}
Large, high-dimensional data sets are now regularly obtained in nearly every  field of research and business. Despite the high dimensions of the data, it is typically assumed that the data can be accurately described using a much smaller set of latent variables. Dimensionality reduction methods aim to find these latent variables and are commonly a key step in the data analysis pipeline. For example, dimensionality reduction is often performed as a preprocessing step as many downstream methods of analysis suffer in performance and accuracy from the `curse of dimensionality'. Principal components analysis (PCA)~\cite{pearson1901,Hotelling1933} is a popular choice for preprocessing because of its simplicity and computational speed. Nonnegative matrix factorization (NMF)~\cite{Lee1999} and classical multi-dimensional scaling (MDS)~\cite{KruskalWish1978} are also commonly used. However, these three methods assume a linear model on the data and thus  are inefficient at representing nonlinear relationships in the latent space. Therefore, much work has focused on nonlinear dimensionality reduction methods  including local linear embedding (LLE)  \cite{Roweis2323}, Isomap \cite{Tenenbaum2319},  diffusion maps~\cite{coifman2006,nadler2006diffusion}, and autoencoders~\cite{hinton1994autoencoders}.

Recent dimensionality reduction methods have focused on reducing dimensions for data visualization. Data visualization is an important aspect of exploratory data analysis for aiding humans in developing an understanding of the underlying structure within the data, which can enable hypothesis generation and data interpretation. Recent, successful visualization methods include  t-SNE (t-Distributed Stochastic Neighbor Embedding) \cite{Maaten2008}, UMAP (Uniform Manifold Approximation and Projection) \cite{lel2018umap}, and PHATE (Potential of Heat-diffusion for Affinity-based Transition Embedding) \cite{Moon2019}. These methods have found applications in computational biology~\cite{amir2013visne,macosko2015highly,gopalakrishnan2018gut,becht2019dimensionality,zhao2020single,karthaus2020regenerative,baccin2019combined},   visualizing text~\cite{chen2011extracting},  time series~\cite{duque2019}, and  the internal nodes in neural networks~\cite{horoi2020low,gigante2019visualizing}.

However, all of these dimensionality reduction and visualization methods are  unsupervised, where class labels are unavailable or ignored. Supervised adaptations have been created for some of these methods. The majority of these \cite{bair2006, yu2006, ZHANG2009, Li2018, Zhang2016, Jia2019, vlachos2002, Ribeiro2008} are used for classification or regression preprocessing, and a few are used for visualization \cite{vlachos2002, tapson2005, Babaee2016, barshan2011}. Many of these approaches incorporate class labels directly in a loss function, which can create an exaggerated separation between classes and introduce other distortions in the visualization. Additionally, some of these adaptations are limited to classification and not easily adapted to regression problems where the labels are not categorical. Finally, these approaches are often sensitive to parameter tuning, which may be difficult to perform without a specific measure of visualization quality. 

In this paper we introduce a new approach to supervised visualization based on random forests \cite{rf} and diffusion-based dimensionality reduction~\cite{coifman2006,nadler2006diffusion}. In particular, we leverage advances made by PHATE~\cite{Moon2019} in extracting the information learned through a diffusion process for visualization. Hence we name our visualization technique RF-PHATE. The approach incorporates variable importance as measured from the trained random forests and gives a noise-resilient visual of the predictor space which may be used in the context of data exploration with known labels. We show empirically that RF-PHATE retains both the local and global structure of the data while emphasizing the important variables in the visualization. We also show that RF-PHATE is robust to noise and parameter tuning.

%ROAD MAP?

%In section \ref{section:Background}, we discuss related works in supervised dimensionality reduction, outlining methodologies, uses, and relation to this work, as well as random forests and their proximities. We then describe our proposed algorithm in section \ref{section:rf-phate} and show results in section \ref{section:results}.

% \section{Preliminaries} \label{section:Background}

\section{Related work on supervised dimensionality reduction}\label{section:relatedwork}
Supervised dimensionality reduction methods can generally be categorized into three major groups: PCA, NMF, and manifold learning~\cite{Chao2019}. The original implementation of supervised PCA (SPCA) selects a subset of features that are most highly correlated with the labels. PCA is applied to these features to generate a new feature space to be used in the regression problem \cite{bair2006}. The primary motivation for this approach was to handle problems where the number of features exceeds the number of observations ($p > n$) \cite{bair2006}. Another variation of SPCA was introduced in \cite{yu2006}, which may be kernelized and is capable of handling missing values. SPCA was applied for data visualization in \cite{barshan2011}. However, SPCA is linear and does not therefore accurately capture non-linear relationships in the data structure in low dimensions. In addition, the number of components selected is usually determined based on global variance explained, which does not adequately capture local relationships~\cite{yu2006}.

NMF \cite{Lee1999} seeks to decompose a data matrix $\mathbf{X}_{n\times p}$ into two, non-negative, matrices $\mathbf{U}_{n\times d}$ and $\mathbf{V}_{d\times p}$, by minimizing the Frobenius norm of the difference between $\mathbf{X}$ and $\mathbf{UV}$ while restricting the entries of $\mathbf{U}$ and $\mathbf{V}$ to be nonnegative.  The rows of $\mathbf{V}$ can be regarded as basis vectors while the columns of $\mathbf{U}$ form the axes of the lower-dimensional space \cite{Tsuge2001}. A number of supervised and semi-supervised modifications to NMF (SNMF or SSNMF) have been proposed, such as constrained NMF \cite{Liu2012}, structured NMF \cite{Li2018}, and NMF for constrained clustering \cite{Zhang2016}. Most of these approaches use the labels in a regularization term in the optimization problem. \citet{Jia2019} proposed a semi-supervised NMF with both similarity and dissimilarity regularization terms. In~\cite{Liu2008,tapson2005,Babaee2016} NMF has also been used for low-dimensional visualization. However, it is still a linear method and therefore does not capture the intrinsic geometric structure of nonlinear data \cite{Cai2008}. In addition, supervised versions of NMF tend to accentuate class differences in clusters, providing inflated separation between groups, and tight clustering within groups.

Manifold-based dimensionality reduction methods assume that the high-dimensional data lie on a low-dimensional manifold.  Examples of manifold-based algorithms include Isomap \cite{Tenenbaum2319}, UMAP \cite{lel2018umap}, Diffusion Maps \cite{coifman2006}, t-SNE \cite{vanDerMaaten2008}, LLE \cite{Roweis2323}, Laplacian Eigenmaps \cite{belkin2003}, and PHATE \cite{Moon2019}. Supervised versions of some of these methods have been proposed. WeightedIso, Iso+Ada~\cite{vlachos2002}, and Enhanced-supervised Isomap (ES-Isomap)~\cite{Ribeiro2008} are supervised versions of Isomap, which estimates the geodesic distances between points. These supervised variations use the class labels to modify the distance metric to accentuate closeness in like-class samples while emphasizing different classes. Supervised versions of LLE (SLLE) have also been proposed that modify the dissimilarity metric using the class information~\cite{ridder2003,ZHANG2009}.  However, like LLE, SLLE is sensitive to parameters. It was also shown in \cite{vlachos2002, Ribeiro2008,polito2002} that LLE and Isomap are subject to ``overclustering'' unless the data are comprised of a single, well-clustered sample.

\section{Supervised neighborhoods via random forest proximities}
\label{RandomForests}

Manifold learning techniques are typically based on the construction of a kernel that computes some measure of pairwise local similarity between data points. For example, classical diffusion maps applies a Gaussian kernel with a fixed bandwidth to pairwise Euclidean distances~\cite{coifman2006} while PHATE applies the $\alpha$-decay kernel with an adaptive $k$-nearest neighbor (NN) bandwidth~\cite{Moon2019}. Isomap similarly builds a $k$-NN graph to measure similarity between points~\cite{Tenenbaum2319}. Trees are also commonly used to construct these kernel measures such as the use of $k$-d trees for computing $k$-NN graphs~\cite{friedman1977kd}. Long range relationships between points are then learned by chaining together the local similarities, for example by diffusion~\cite{coifman2006,Moon2019} or shortest path algorithms~\cite{Tenenbaum2319}.

To create supervised versions of manifold learning techniques, the label information can be incorporated in the kernel construction. We propose to do this using random forests.  Random forests \cite{rf} are a tree-based ensemble method commonly used for classification and regression. They are widely considered one of the best out-of-the-box supervised learning techniques as they often achieve very good results with relatively little tuning~\cite{Cutler2012}. In addition to classification and regression, random forests have a number of additional uses, such as variable selection, imputation, outlier detection, and unsupervised learning \cite{Cutler2012}. 

Random forests were created as an extension to Leo Breiman's bagging idea \cite{Breiman1996}, which uses bootstrap samples (sampling with replacement) to train base learners (decision trees in this case). The out-of-sample data is said to be ``out of bag'' \cite{rf}, which is often used as a test set to estimate the generalization error of the model as well as to assess variable importance \cite{Cutler2012}.  The aggregate results of the base learners are combined to make a final prediction \cite{Breiman1996}. While individual decision trees tend to be sensitive to the data,  random forests are very stable due to the two-part randomness of the algorithm (boostrapping and random variable selection), which ensures low correlation among the base learners~\cite{segal2011}. Random forests are also robust to noise as splits are not likely to be determined by noise variables when meaningful variables are considered at a given node. Overall, as the number of trees increases, meaningful variables are more likely to be shown as important while noise variables will show less importance, averaging over all trees.

\begin{definition}
\label{def:prox}
Consider two out-of-bag observations from the training data $x_i$ and $x_j$.  The random forest proximity between these points $K_{RF}(x_i,x_j)$ is defined as the proportion of trees in the trained random forest for which the observations $x_i$ and $x_j$ share the same terminal (leaf) node  \cite{rfsd:Online}.
\end{definition}

From Definition~\ref{def:prox}, it is clear that two observations that always end in the same terminal node will have a proximity of 1, while those that never fall into the same terminal node will have a proximity of 0. The proximities applied to all pairs of training points form a symmetric, positive definite kernel matrix with ones along the diagonal. The random forest proximities provide a measure of how similar observations are in the predictor space and naturally take into account the weight of variable importance relative to the training task.  That is to say, two observations may be near one another in the predictor space due to having similar values in  relatively few, important variables, but may otherwise be far apart in a Euclidean space \cite{Cutler2012}. Thus using the proximities as inputs to visualization methods will produce visualizations that emphasize the variables important to the supervised task while largely ignoring the irrelevant variables \cite{rfadaptiveknn}.

We propose to use the random forest proximities as the kernel for RF-PHATE because of their desirable properties described above. Random forest proximities have been used previously for visualization by applying MDS~\cite{Cutler2012} or graph-based methods~\cite{golino2014}. However, applying MDS directly to the proximities loses much of the geometric structure of the data (see figure \ref{fig:proxmethods} in appendix \ref{sec:prox}), while force-directed layout methods (such as the Fruchterman-Reingold algorithm \cite{frucht1991} used in \cite{golino2014})  do not scale well with large data sets \cite{GAJER20043}. In contrast, our approach preserves the geometric structure (both locally and globally) due to our use of diffusion and information distances and scales well to both small and large data sets.

\section{RF-PHATE}

Kernels typically contain mostly local information about the data points used for training rather than global. This includes the random forest proximities. Consider a classification task with two points $x$ and $y$ that have the same class label. Suppose $x$ and $y$ are similar to each other in the variables that are important for classification. Then $x$ and $y$ will have a high proximity as measured by a trained random forest as they are likely to end up in the same terminal node of each tree. This is true even if the points are highly dissimilar in variables that are not important for the classification. Now, suppose $x$ and $y$ have the same class label but are dissimilar in one or more variables that are relevant for classification. Then $x$ and $y$ will have a lower proximity between them compared to the first scenario as it is less likely they will end up in the same terminal nodes. However, the proximities between sufficiently dissimilar points will all be zero, thus obscuring some of the long-range relationships in the data. Therefore, directly embedding the proximities leads to global distortion. 

A robust way of learning global relations from local proximities is via diffusion (see~\cite{coifman2006,nadler2006diffusion, Moon2019}). Because of recent successes in adapted, diffusion-based visualizations (see~\cite{Moon2019, duque2019visualizing,gigante2019visualizing,horoi2020low}), we create a new, diffusion-based, supervised visualization tool, called RF-PHATE, which combines the locally-adaptive random forest kernel with recent advances in manifold learning; all of the steps are provided in Algorithm~\ref{rfPhateAlg}. Random forests can be viewed as a $k$-nearest neighbor (NN) classifier with an adaptive metric~\cite{rfadaptiveknn}. The proximities produced by the random forest can therefore be viewed as locally-adapted affinities in the predictor space, where local variable importance is assessed using a measure of class purity at each node, making the affinities insensitive to data density. This further motivates the use of random forest proximities to encode local structure in data that is dependent on the label space. 

\begin{wrapfigure}[13]{R}{0.5\linewidth}
\vspace{-12pt}
\centering
\scriptsize
\begin{algorithm}[H]
\SetAlgoLined
 \textbf{Input}: Data matrix $X (n\times p)$, class labels $y$, output dimension $m$\\
 \textbf{Output}: $Y_m$, an $m$-dimensional ($m < p$) embedding
 \begin{enumerate}
     \item Compute the random forest proximities, $K$
     \item $P\leftarrow$ row-normalize $K$ to form the diffusion operator
     \item $t\leftarrow$ compute time scale via Von Neumann Entropy
     \item $P^t\leftarrow$ diffuse $P$ for $t$ time steps
     \item Compute potential representations $U_t \leftarrow -\log(P^t)$
     \item Compute potential distances $\mathscr{D}_t$ from $U_t$
     \item $Y_{\text {class}} \leftarrow$ apply classical MDS to $\mathscr{D}_{t}$
     \item $Y_m \leftarrow$ apply metric MDS to $\mathscr{D}_t$ with initialization $Y_\text{class}$
 \end{enumerate}
\caption{The RF-PHATE algorithm  \label{rfPhateAlg}}
\end{algorithm}
\end{wrapfigure}

Diffusion maps constructs a graph from local similarities to learn the global data geometry. Typically, the graph is constructed by applying a kernel function (e.g. Gaussian) to pairwise Euclidean distances using a fixed kernel bandwidth resulting in an $N\times N$ kernel matrix $K$, where $N$ is the number of points we wish to embed, giving a notion of similarity between data points. Instead of applying a kernel function to Euclidean distances, our method uses the random forest proximity (local-affinity) matrix. The local affinity matrix is then row-sum normalized to create a Markov transition matrix or diffusion operator, $P$, which is used to capture global relationships. An entry, $j$, of row $i$ in $P$ represents the probability of transitioning or ``walking'' from point $i$ to point $j$ in a single step of a random walk on a graph constructed from the affinity matrix $K$. Thus the probability of transitioning from point $x$ to point $y$ on this graph in a single time step is high if the distance between them is small. This prevents an exiting of the intrinsic manifold of the data. $P$ is raised to the power of $t$ to simulate $t$ diffusion steps. This has the effect of denoising as transitions with low probabilities are filtered out, while transitions with high probability retain their importance. In diffusion maps, the local and global information learned in $P^t$ is often extracted into lower dimensions using eigendecomposition.

Diffusion maps has several weaknesses that prevent it from being effective at visualization in most settings. First, in many applications a fixed bandwidth for all points is not appropriate as the data may not be sampled uniformly. Second, choosing a good time scale $t$ for the diffusion process is difficult and largely overlooked in classical diffusion maps. A small value of $t$ can lead to insufficient denoising and an overemphasis on local structure. In contrast, a large value of $t$ can lead to oversmoothing and an overemphasis on global structure.  Third, the eigendecomposition in diffusion maps tends to place the information learned in $P^t$ into different dimensions~\cite{Moon2019,haghverdi2016diffusion}, which is not amenable for visualization. 

Since the random forest proximity matrix, $K$, is comprised of  locally-adaptive affinities, it does not rely on uniformly-sampled data, overcoming this weakness in diffusion maps. The von Neumann Entropy (VNE) of the diffused operator provides a good choice of $t$ for visualization. The VNE of the diffused operator $P^t$ is the Shannon entropy of the normalized eigenvalues of $P^t$. Since the entropy of a discrete random variable is maximized with a uniform distribution, the VNE is a soft proxy for the number of significant eigenvalues of $P^t$. The more significant eigenvalues there are, the closer the normalized eigenvalues are to a uniform distribution, and the higher the VNE. Since $P$ is a probability transition matrix with a stationary distribution, the VNE converges to zero as $t\rightarrow \infty$. The rate of decay of the VNE as $t$ increases is used to select $t$. Typically, $t$ is chosen to be around the transition from rapid to slow decay in the VNE as this is considered to be point in the diffusion process where noise has been eliminated and oversmoothing begins~\cite{Moon2019}. We show empirically (see appendix \ref{section:paramselection}) that diffusion over proximities is not sensitive to the choice of $t$ around the VNE-selected value for local variable preservation.

To counter the third weakness of diffusion maps, we apply an information distance to the powered diffusion operator $P^t$ to create the potential distance~\cite{Moon2019}. The potential distance is calculated by log- or square-root-transforming the powered diffusion operator and then calculating the Euclidean distance between rows, although other information distances can also be used~\cite{duque2019visualizing}. The potential distance is sensitive to differences in both the tails and the more dense regions of the diffused probabilities, resulting in a distance which preserves both local and global relationships. These distances are then embedded into low-dimensions using metric MDS with classical MDS as an initialization. This extracts the information in low dimensions for better visualization.

\section{Experimental Results} \label{section:results}
To evaluate our supervised visualization method, we apply RF-PHATE to multiple data sets and compare it to other methods described in Sections~\ref{section:intro} and \ref{section:relatedwork}. Full details on the used data sets, computational environment, method implementations, and parameter tuning can be found in the appendices. We first demonstrate qualitatively how RF-PHATE can be used in exploratory data analysis to visualize variable importance with respect to the supervised task on both low- and high-dimensional data sets. We then quantitatively establish this capability compared to other supervised dimensionality reduction methods. Finally, we show that RF-PHATE is able to accurately visualize data with noisy dimensions and outperforms both supervised and unsupervised dimensionality reduction methods.

\subsection{Visualizing Variable Importance}

\begin{figure}[!htb]
    \centering
    \includegraphics[width = 0.9\textwidth]{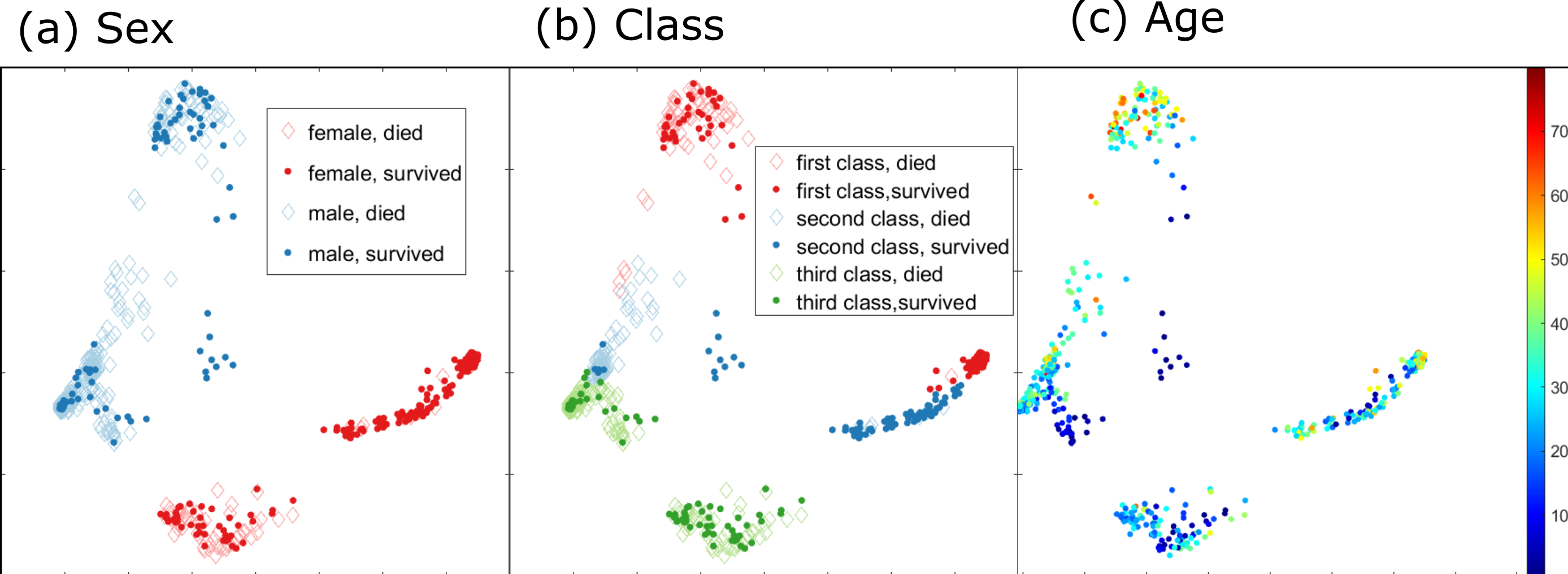}
    \caption{RF-PHATE on the \textit{Titanic} data set colored by sex, class, and age.  Passenger deaths are denoted by light-colored diamonds while survivors are marked by dark-colored dots in (a) and (b).  Sex and class were the top two important variables. The visualization  shows clear groups and trends within the data.}
    \label{fig:titanic}
\end{figure}

We demonstrate the ability of RF-PHATE to capture local variable importance in a low-dimensional representation of the popular \textit{Titanic} data set~\cite{UCI2019}, which has 12 variables and 891 observations.  The variables \texttt{Name} and \texttt{Ticket} were removed as they are nearly unique to each observation.  In addition, observations with missing values were removed. The proximities were created using the \texttt{randomForest} \cite{randomForest} package in \texttt{R} \cite{R}, with the variable \texttt{Survived} as the response. The forests' classification accuracy was 81.46\% using 5000 trees  and otherwise default parameters. Variable importance was computed using the forest's built-in \texttt{importance} measure (mean decrease in accuracy). To compute the importance for the $m^{th}$ variable using a random forest, the error rate (for classification) or MSE (for regression) is computed using the out-of-bag (oob) observations. The values of $m$ for the oob data are randomly permuted and the error rate (or MSE) is computed using the permuted data.  The importance is the difference between the oob error rate and the permuted oob error rate \cite{Cutler2012}. For the \texttt{Titanic} data set, the top two important variables were \texttt{Sex} and \texttt{{Class}}. Figure~\ref{fig:titanic} shows the results of applying RF-PHATE to this data. Instead of creating distinct clusters based solely on the response, RF-PHATE arranges points and clusters by variable importance. The right-most cluster includes mostly survivors and contains female first- and second-class passengers. Most passengers in the left-most cluster did not survive. This cluster contains male, second- and third-class passengers. The small cluster in the center is entirely composed of second-class, male children who all survived. At least some of the variation within clusters appears to be driven by age. This demonstrates how RF-PHATE can be used in data exploration by  visually identifying groupings and trends within the variables important for the supervised task. 

\subsection{Visualization of Higher-Dimensional Data}

\begin{figure}[!ht]
    \centering
    \includegraphics[width = \textwidth]{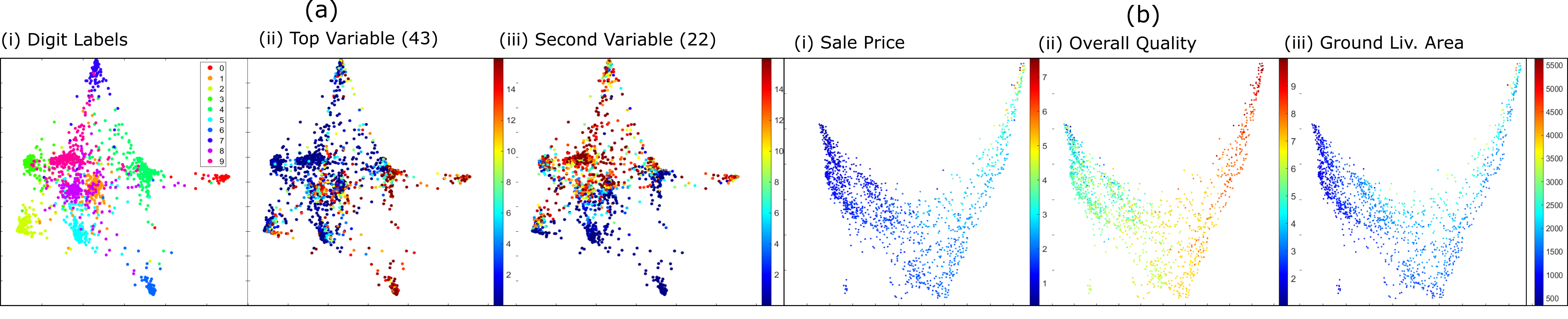}
    \caption{(a) RF-PHATE on the \textit{Optical Digits} data set~\cite{xu1992digits, UCI2019}.  Some numbers (0, 6) are easily separated, while others are more difficult to classify, such as 1 and 9.  The color-bars in (a)ii and (a)iii indicate the count of pixels made up by $4\times4$ non-overlapping blocks. Numbers 0, 3, and 6 have higher counts in the pixels which make up the $43^{rd}$ block (the top variable) as can be seen in (a)ii.  (b) RF-PHATE on the \textit{Ames Housing} data set~\cite{cock2011ames}. By comparing (b)i, (b)ii, and (b)iii it is easy to see the relationship between higher sale prices with overall house quality and ground-floor living area.}
    \label{fig:optdigits}
\end{figure}

To demonstrate the application of RF-PHATE to high dimensional data we apply it to the Optical Recognition of Handwritten Digits data from~\cite{xu1992digits, UCI2019} and Ames, Iowa, house pricing data from~\cite{cock2011ames}, with classification and regression supervision, respectively. The Optical Digits data contains 5620 observations of handwritten numerical digits  recorded in $32\times32$ bit images. The images are divided into $4\times4$ blocks resulting in 64 dimensional arrays \cite{xu1992digits, UCI2019}. The random forest accuracy for the data set was 82.7\%. The variables in this data set are not as straightforward to analyze as they consist of individual pixels in the images, but we can see in Figure~\ref{fig:optdigits}(a) that high values in the top variable (block 43) are important in classifying 0's and 6's, while high values in the second variable provide further separation between 4's, 7's, and 9's from the 3's, 5's, and 6's. On the Ames, Iowa house pricing data set \cite{cock2011ames}, we adapted the random forest in RF-PHATE to perform regression on house prices, resulting in overall quality and ground living area (square footage) being designated as the most important variables, as emphasized by the RF-PHATE visualization in Figure~\ref{fig:optdigits}(b).

\subsection{Variable Importance Preservation}
\label{sub:numbers}
We now numerically assess the quality of the embeddings provided by RF-PHATE. To be useful in supervised exploratory data analysis, the embedding should preserve local and global structure in the most important variables for the supervised task. To numerically assess this structure preservation, we applied a $k$-nearest neighbor classifier or regressor with the embedding dimensions as inputs (i.e. features) and the important variables as the response output.  The key idea is that if the embedding preserves the local and global structure in the important variables, then the embedding dimensions should be sufficient to predict the important variables. 

For classification, the mean prediction error, and for regression, the root mean-squared error was used as the criterion, applying 10-fold cross-validation. We compared RF-PHATE with six supervised methods: Enhanced Supervised Isomap (ESIso) \cite{Ribeiro2008}, Kernalized Supervised PCA (KSPCA) \cite{yu2006}, SPCA \cite{yu2006}, Enhanced Supervised LLE (ESLLE) \cite{ZHANG2009}, SNMF \cite{Jia2019}, and Partial Least Squares Discriminant Analysis (PLSDA) \cite{wold1984}. 

\begin{table}
\footnotesize
\centering
\caption{Numerical comparison of RF-PHATE with other supervised methods. The top two to  four variables for  classification (Variable \textit{Titanic} and  \textit{Optical Digits}) or regression (\textit{Ames Housing}) were used as the response, using the low-dimensional embeddings (2 or 3) from 6 supervised dimensionality reduction algorithms. For categorical variables (Sex, Class) the classification error was used; for continuous variables (all of the remaining), the root mean-squared error (RMSE) was used. Therefore, lower is better. ESIso, ESLLE, and SNMF have not been adapted for regression, hence the missing values in the Ames dataset (N/A). RF-PHATE is either first or second at each task. \label{tab:varErr}}
\begin{tabular}{|c||c|c|c|c|c|c|c|c|}
\hline
    Data Set & Variable (dims) & RF-PHATE	&	ESIso	&	KSPCA	&	SPCA	&	ESLLE	&	SNMF	&	PLSDA	\\
\hline
Titanic & Sex (2) & \underline{\textbf{0.0000}}	&	\textbf{0.0028}	&	0.0084	&	0.0056	&	0.0070	&	0.1419	&	\textbf{0.0028}	\\
Titanic & Sex (3) & \underline{\textbf{0.0000}}	&	\textbf{0.0028}	&	0.0084	&	\textbf{0.0028}	&	0.0042	&	0.1489	&	0.0042	\\
\hline
Titanic & Class (2) & \underline{\textbf{0.0154}}	&	0.1657	&	0.1025	&	0.0688	&	0.0702	&	0.3174	&	\textbf{0.0197}	\\
Titanic & Class (3) & \underline{\textbf{0.0112}}	&	0.0492	&	0.1011	&	0.0281	&	0.0969	&	0.3048	&	\textbf{0.0239}	\\
\hline
Optical & Top Var (2)	&	\textbf{3.612}	&	\underline{\textbf{3.554}}	&	5.478	&	5.237	&	5.910	&	6.659	&	5.842	\\
Optical & Top Var (3)	&	\underline{\textbf{3.473}}	&	\textbf{3.484}	&	4.013	&	4.922	&	5.887	&	6.659	&	5.407	\\
\hline
Optical & Second Var (2)	&	\underline{\textbf{3.425}}	&	\textbf{3.869}	&	4.101	&	4.862	&	5.441	&	6.540	&	5.730	\\
Optical & Second Var (3)	&	\underline{\textbf{3.361}}	&	\textbf{3.851}	&	3.902	&	4.007	&	5.194	&	6.540	&	5.060	\\
\hline
Optical & Third Var (2)	&	\textbf{4.147}	&	\underline{\textbf{3.997}}	&	5.107	&	4.416	&	5.839	&	7.147	&	6.217	\\
Optical & Third Var (3)	&	\underline{\textbf{3.766}}	&	\textbf{3.810}	&	4.608	&	4.039	&	5.810	&	7.147	&	5.968	\\
\hline
Optical & Fourth Var (2)	&	\underline{\textbf{4.108}}	&	\textbf{4.118}	&	5.373	&	5.687	&	5.965	&	6.161	&	5.985	\\
Optical & Fourth Var (3)	&	\underline{\textbf{4.001}}	&	4.037	&	\textbf{4.035}	&	4.949	&	5.867	&	6.161	&	5.320\\
\hline
    Ames & Quality (2)	&	\underline{\textbf{0.644}}	&	N/A	&	0.815	&	\textbf{0.764}	&	N/A	&	N/A	&	0.767	\\
    Ames & Quality (3)	&	\underline{\textbf{0.631}}	&	N/A	&	0.788	&	0.759&	N/A	&	N/A	&	\textbf{0.736}\\
\hline
    Ames & GrLvArea (2)	&	\underline{\textbf{263.75}}	&	N/A	&	\textbf{402.08}	&	413.36	&	N/A	&	N/A	&	415.90	\\
    Ames & GrLvArea (3)	&	\underline{\textbf{224.57}}	&	N/A	&	393.64	&	411.53	&	N/A	&	N/A	&	\textbf{293.18}	\\
\hline
\end{tabular}
\end{table}

Table~\ref{tab:varErr} shows the results for the Titanic, Optical Digits, and Ames Housing data sets, respectively. For the Titanic data set, RF-PHATE performs the best out of all methods when predicting either sex or class (the two most important variables) using either a 2 or 3 dimensional embedding. For the Optical Digits, RF-PHATE is either the best or second-best method at preserving the relevant structure when using 2 or 3 dimensions. ESIso is generally in second and is sometimes significantly below RF-PHATE in terms of quality. We performed similar comparisons on other data sets. We also note that RF-PHATE continues to do well when embedding with higher dimensions.  See appendices \ref{section:data} and \ref{section:varregress}.

\subsection{Supervised Visualization in the Presence of Noisy Variables}

\begin{figure}[!htb]
    \centering
    \includegraphics[width = 0.9\textwidth]{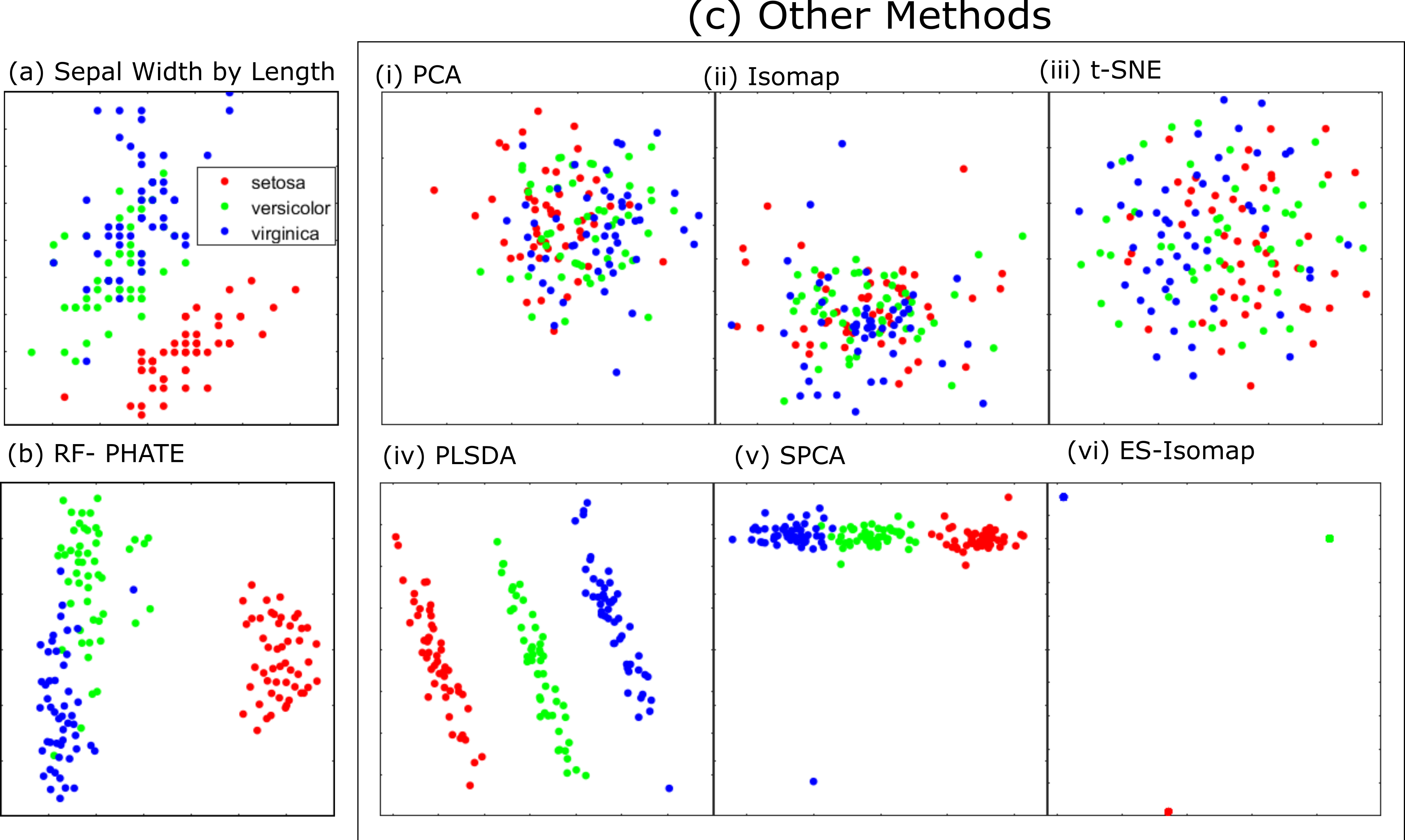}
    \caption{Iris data set \cite{anderson1936iris, UCI2019} with additional 1000 Gaussian noise variables using various dimensionality reduction techniques.  The ground truth (without additional noise) is shown in (a). RF-PHATE (b) effectively denoises the data and produces a comparable visualization.  It is seen that unsupervised methods (c)[(i), (ii), and (iii)] do not perform in this scenario. Other supervised methods (c)[(iv), (v), and (vi)] separate the data into classes but lose some of the general data structure.}
    \label{fig:iris}
\end{figure}

 We now consider the problem where only a few variables are relevant for the supervised task and all other variables contain noise. This is a case where unsupervised visualization methods are very likely to fail, demonstrating the utility of supervised visualization methods.  To simulate this, we took the Iris data set \cite{anderson1936iris, UCI2019} and added 1000 dimensions generated from a Gaussian random variable with means uniformly generated between -1 and 1 and with a variance of 1. The simulation was repeated 10 times. This also creates a case with few data points as the Iris data set has only 150 samples. The visualizations are shown in Figure~\ref{fig:iris}. We note that the unsupervised methods (PCA, Isomap, and t-SNE) perform poorly in this setting as their visualizations consist of a single cloud of points without any distinguishable features.

In contrast, the supervised visualization methods are able to reveal some structure by using the class labels. RF-PHATE (Figure \ref{fig:iris}(b)) separates the setosa class from the other two classes while the versicolor and virginica classes are shown with some overlap.  Furthermore, the setosa class is shown to be approximately the same distance from each of the other two classes. This overall structure is consistent with the properties of the sepal measurements in the Iris data set as seen in figure~\ref{fig:iris}(a) in appendix \ref{section:data}. 

PLSDA and ES-Isomap (Figures~\ref{fig:iris}(c)iv and vi, respectively) separate the classes well. However, these methods do not show the overlap between the versicolor and virginica species. Also, the ES-Isomap visualization shows the data points for each class collapsed onto a single point, completely eliminating any structure that may be present within the classes.  This suggests that these methods do not preserve global relationships between classes well.

Supervised PCA (Figure \ref{fig:iris}(c)v)  separates the setosa class from the other two classes while the versicolor and virginica species share some overlap.  The setosa species is shown in the visualization to be much closer to the versicolor class than the virginica species. While this is not consistent with the global structure of the sepal measurements (Figure~\ref{fig:iris}(a)), it is consistent with the structure of the petal measurements (see figure \ref{fig:irisPairs}). However, SPCA does not show a large separation between the setosa and versicolor classes, suggesting that SPCA may not always accurately reflect the magnitude of class separations. 

These results are corroborated numerically in Table~\ref{tab:iris}. Here we performed a similar experiment as in Section~\ref{sub:numbers} where we performed regression on the embeddings to predict each of the 4 variables in the data set. Again, RF-PHATE outperforms the other methods in all settings. While this is a simple data set, its simplicity allows us to evaluate qualitatively and quantitatively the performance of the various methods, and thus reflects a generalizable trend that will be studied further in future work. See the appendix \ref{section:varregress} for experiments with higher dimensions.

\begin{table}
\scriptsize
\centering
\caption{Variable regression on \textit{Iris} with 1000 additional random noise variables.  The low-dimensional embeddings of seven supervised dimensionality reduction methods were used as features to regress on the original iris data set variables. The noise variables were simulated ten times and the results were averaged. The standard deviation is also reported. RF-PHATE outperforms all other methods. \label{tab:iris}}
\begin{tabular}{|c|c|c|c|c|c|c|c|}
\hline
    Var (dims)	&	RF-PHATE		&	ESIso		&	KSPCA		&	SPCA		&	ESLLE		&	SNMF		&	PLSDA		\\

    \hline
     petal len. (2)	&	\underline{\textbf{0.330}}	$\pm$ \underline{\textbf{0.01}}	&	0.348	$\pm$ 0.01	&	\textbf{0.340}	$\pm$ \textbf{0.01}	&	0.381	$\pm$ 0.01	&	0.349	$\pm$ 0.01	&	1.475	$\pm$ 0.03	&	0.387	$\pm$ 0.01	\\
    petal len. (3)	&	\underline{\textbf{0.334}}	$\pm$ \underline{\textbf{0.01}}	&	\textbf{0.353}	$\pm$ \textbf{0.01}	&	0.360	$\pm$ 0.01	&	0.400	$\pm$ 0.01	&	\textbf{0.353}	$\pm$ \textbf{0.01}	&	2.233	$\pm$ 0.04	&	0.363	$\pm$ 0.01	\\

    \hline	
    
    petal wid. (2)	&	\underline{\textbf{0.291}}	$\pm$ \underline{\textbf{0.01}}	&	0.298	$\pm$ 0.01	&	\textbf{0.297}	$\pm$ \textbf{0.01}	&	0.330	$\pm$ 0.01	&	0.304	$\pm$ 0.01	&	1.297	$\pm$ 0.02	&	0.330	$\pm$ 0.01	\\
    petal wid. (3)	&	\underline{\textbf{0.292}}	$\pm$ \underline{\textbf{0.01}}	&	\textbf{0.305}	$\pm$ \textbf{0.01}	&	0.312	$\pm$ 0.01	&	0.345	$\pm$ 0.01	&	0.308	$\pm$ 0.01	&	1.905	$\pm$ 0.03	&	0.308	$\pm$ 0.01	\\

    \hline
    
    sepal len. (2)	&	\underline{\textbf{0.459}}	$\pm$ \underline{\textbf{0.01}}	&	0.522	$\pm$ 0.02	&	\textbf{0.514}	$\pm$ \textbf{0.02}	&	\textbf{0.514}	$\pm$ \textbf{0.02}	&	0.528	$\pm$ 0.02	&	0.885	$\pm$ 0.02	&	\textbf{0.514}	$\pm$ \textbf{0.02}	\\
    sepal len. (3)	&	\underline{\textbf{0.455}}	$\pm$ \underline{\textbf{0.01}}	&	0.536	$\pm$ 0.02	&	0.540	$\pm$ 0.02	&	0.520	$\pm$ 0.02	&	0.534	$\pm$ 0.02	&	1.272	$\pm$ 0.02	&	\textbf{0.516}	$\pm$ \textbf{0.02}	\\

    \hline
    
    sepal wid. (2)	&	\underline{\textbf{0.320}}	$\pm$ \underline{\textbf{0.01}}	&	\textbf{0.348}	$\pm$ \textbf{0.01}	&	0.351	$\pm$ 0.01	&	0.357	$\pm$ 0.01	&	0.349	$\pm$ 0.01	&	0.462	$\pm$ 0.01	&	0.357	$\pm$ 0.01	\\
    sepal wid. (3)	&	\underline{\textbf{0.318}}	$\pm$ \underline{\textbf{0.01}}	&	\textbf{0.354}	$\pm$ \textbf{0.01}	&	0.358	$\pm$ 0.01	&	0.356	$\pm$ 0.01	&	\textbf{0.354}	$\pm$ \textbf{0.01}	&	0.501	$\pm$ 0.02	&	0.357	$\pm$ 0.01	\\
\hline

\end{tabular}
\end{table}

\section{Conclusion}

Large, high-dimensional data sets are commonplace and can be difficult to analyze and interpret.  Supervised dimensionality reduction algorithms are used for preprocessing and visualizing  high-dimensional data sets when labels are present. Current methods for supervised visualization either over-emphasize class separation or ignore the data's intrinsic geometric structure altogether. In contrast, RF-PHATE incorporates labels directly in its kernel matrix using random forests' out-of-bag samples, thereby robustly learning local relationships without overstressing class separation. The global structure is discovered by means of diffusion and then visualized by embedding an information distance. RF-PHATE outperforms other supervised dimensionality reduction algorithms, is noise-resilient, robust to parameter tuning, and handles both continuous and discrete labels.

\section*{Broader Impact}

Data exploration plays a role in virtually all fields of science and technology nowadays. In this era of big data, there is an ever-increasing need for means to make sense of the information overload we face daily. In instances where labeled data is present, RF-PHATE has the potential to visually unlock patterns and signals in excessively noisy data. This holds the potential to lead to new discoveries in social, medical, natural, and business settings. RF-PHATE's implementation allows enables identification of trends or clusters based on important variables. These clusters are inherently meaningful and not subject to arbitrary subjective interpretation that is often encountered in an unsupervised setting. This, in turn, may lead to the discovery of meaningful relationships not only between the variables and classes, but also between the variables themselves. We do note that like many other data processing tools, data visualization is not immune to misuse, and may be used as a means of deceit when not presented properly. However, this work is computational in nature and addresses fundamental development in data science, agnostic to specific application. As such, by itself, it is not expected to raise ethical concerns nor to have adverse effects on society.

\bibliographystyle{unsrtnat}
\bibliography{main}

\begin{thebibliography}{67}
\providecommand{\natexlab}[1]{#1}
\providecommand{\url}[1]{\texttt{#1}}
\expandafter\ifx\csname urlstyle\endcsname\relax
  \providecommand{\doi}[1]{doi: #1}\else
  \providecommand{\doi}{doi: \begingroup \urlstyle{rm}\Url}\fi

\bibitem[Pearson(1901)]{pearson1901}
Karl Pearson.
\newblock {LIII. On} lines and planes of closest fit to systems of points in
  space.
\newblock \emph{The London, Edinburgh, and Dublin Philosophical Magazine and
  Journal of Science}, 2\penalty0 (11):\penalty0 559--572, 1901.

\bibitem[Hotelling(1933)]{Hotelling1933}
Harold Hotelling.
\newblock Analysis of a complex of statistical variables into principal
  components.
\newblock \emph{Journal of Educational Psychology}, 1933.

\bibitem[Lee and Seung(1999)]{Lee1999}
Daniel~D. Lee and H.~Sebastian Seung.
\newblock Learning the parts of objects by non-negative matrix factorization.
\newblock \emph{Nature}, 401\penalty0 (6755):\penalty0 788--791, Oct 1999.
\newblock ISSN 1476-4687.

\bibitem[Kruskal and Wish(1978)]{KruskalWish1978}
J.B. Kruskal and M.~Wish.
\newblock \emph{{Multidimensional Scaling}}.
\newblock Sage Publications, 1978.

\bibitem[Roweis and Saul(2000)]{Roweis2323}
Sam~T. Roweis and Lawrence~K. Saul.
\newblock Nonlinear dimensionality reduction by locally linear embedding.
\newblock \emph{Science}, 290\penalty0 (5500):\penalty0 2323--2326, 2000.
\newblock ISSN 0036-8075.

\bibitem[Tenenbaum et~al.(2000)Tenenbaum, Silva, and Langford]{Tenenbaum2319}
Joshua~B. Tenenbaum, Vin~de Silva, and John~C. Langford.
\newblock A global geometric framework for nonlinear dimensionality reduction.
\newblock \emph{Science}, 290\penalty0 (5500):\penalty0 2319--2323, 2000.
\newblock ISSN 0036-8075.

\bibitem[Coifman and Lafon(2006)]{coifman2006}
Ronald~R. Coifman and Stéphane Lafon.
\newblock Diffusion maps.
\newblock \emph{Applied and Computational Harmonic Analysis}, 21\penalty0
  (1):\penalty0 5 -- 30, 2006.
\newblock ISSN 1063-5203.
\newblock Special Issue: Diffusion Maps and Wavelets.

\bibitem[Nadler et~al.(2006)Nadler, Lafon, Kevrekidis, and
  Coifman]{nadler2006diffusion}
Boaz Nadler, Stephane Lafon, Ioannis Kevrekidis, and Ronald~R Coifman.
\newblock Diffusion maps, spectral clustering and eigenfunctions of
  fokker-planck operators.
\newblock In \emph{Advances in neural information processing systems}, pages
  955--962, 2006.

\bibitem[Hinton and Zemel(1994)]{hinton1994autoencoders}
Geoffrey~E Hinton and Richard~S Zemel.
\newblock Autoencoders, minimum description length and helmholtz free energy.
\newblock In \emph{Advances in neural information processing systems}, pages
  3--10, 1994.

\bibitem[Maaten and Hinton(2008)]{Maaten2008}
Laurens van~der Maaten and Geoffrey Hinton.
\newblock Visualizing data using t-{SNE}.
\newblock \emph{Journal of machine learning research}, 9\penalty0
  (Nov):\penalty0 2579--2605, 2008.

\bibitem[McInnes et~al.(2018)McInnes, Healy, and Melville]{lel2018umap}
Leland McInnes, John Healy, and James Melville.
\newblock {UMAP: U}niform manifold approximation and projection for dimension
  reduction, 2018.

\bibitem[Moon et~al.(2019)Moon, van Dijk, Wang, Gigante, Burkhardt, Chen, Yim,
  Elzen, Hirn, Coifman, Ivanova, Wolf, and Krishnaswamy]{Moon2019}
Kevin~R. Moon, David van Dijk, Zheng Wang, Scott Gigante, Daniel~B. Burkhardt,
  William~S. Chen, Kristina Yim, Antonia van~den Elzen, Matthew~J. Hirn,
  Ronald~R. Coifman, Natalia~B. Ivanova, Guy Wolf, and Smita Krishnaswamy.
\newblock Visualizing structure and transitions in high-dimensional biological
  data.
\newblock \emph{Nature Biotechnology}, 37\penalty0 (12):\penalty0 1482--1492,
  Dec 2019.
\newblock ISSN 1546-1696.

\bibitem[Amir et~al.(2013)Amir, Davis, Tadmor, Simonds, Levine, Bendall,
  Shenfeld, Krishnaswamy, Nolan, and Pe'er]{amir2013visne}
El-ad~David Amir, Kara~L Davis, Michelle~D Tadmor, Erin~F Simonds, Jacob~H
  Levine, Sean~C Bendall, Daniel~K Shenfeld, Smita Krishnaswamy, Garry~P Nolan,
  and Dana Pe'er.
\newblock visne enables visualization of high dimensional single-cell data and
  reveals phenotypic heterogeneity of leukemia.
\newblock \emph{Nature Biotechnology}, 31\penalty0 (6):\penalty0 545, 2013.

\bibitem[Macosko et~al.(2015)Macosko, Basu, Satija, Nemesh, Shekhar, Goldman,
  Tirosh, Bialas, Kamitaki, Martersteck, et~al.]{macosko2015highly}
Evan~Z Macosko, Anindita Basu, Rahul Satija, James Nemesh, Karthik Shekhar,
  Melissa Goldman, Itay Tirosh, Allison~R Bialas, Nolan Kamitaki, Emily~M
  Martersteck, et~al.
\newblock Highly parallel genome-wide expression profiling of individual cells
  using nanoliter droplets.
\newblock \emph{Cell}, 161\penalty0 (5):\penalty0 1202--1214, 2015.

\bibitem[Gopalakrishnan et~al.(2018)Gopalakrishnan, Spencer, Nezi, Reuben,
  Andrews, Karpinets, Prieto, Vicente, Hoffman, Wei,
  et~al.]{gopalakrishnan2018gut}
Vancheswaran Gopalakrishnan, Christine~N Spencer, Luigi Nezi, Alexandre Reuben,
  MC~Andrews, TV~Karpinets, PA~Prieto, D~Vicente, Karen Hoffman, SC~Wei, et~al.
\newblock Gut microbiome modulates response to anti--pd-1 immunotherapy in
  melanoma patients.
\newblock \emph{Science}, 359\penalty0 (6371):\penalty0 97--103, 2018.

\bibitem[Becht et~al.(2019)Becht, McInnes, Healy, Dutertre, Kwok, Ng, Ginhoux,
  and Newell]{becht2019dimensionality}
Etienne Becht, Leland McInnes, John Healy, Charles-Antoine Dutertre,
  Immanuel~WH Kwok, Lai~Guan Ng, Florent Ginhoux, and Evan~W Newell.
\newblock Dimensionality reduction for visualizing single-cell data using umap.
\newblock \emph{Nature Biotechnology}, 37\penalty0 (1):\penalty0 38, 2019.

\bibitem[Zhao et~al.(2020)Zhao, Amodio, Vander~Wyk, Gerritsen, Kumar, van Dijk,
  Moon, Wang, Malawista, Richards, Cahill, Desai, Sivadasan, Venkataswamy,
  Ravi, Fikrig, Kumar, Kleinstein, Krishnaswamy, and
  Montgomery]{zhao2020single}
Yujiao Zhao, Matthew Amodio, Brent Vander~Wyk, Bram Gerritsen, Mahesh~M Kumar,
  David van Dijk, Kevin Moon, Xiaomei Wang, Anna Malawista, Monique~M Richards,
  Megan Cahill, Anita Desai, Jayasree Sivadasan, Manjunatha Venkataswamy,
  Vasanthapuram Ravi, Erol Fikrig, Priti Kumar, Steven Kleinstein, Smita
  Krishnaswamy, and Ruth Montgomery.
\newblock Single cell immune profiling of dengue virus patients reveals intact
  immune responses to {Zika} virus with enrichment of innate immune signatures.
\newblock \emph{PLoS Neglected Tropical Diseases}, 14\penalty0 (3):\penalty0
  e0008112, 2020.

\bibitem[Karthaus et~al.(2020)Karthaus, Hofree, Choi, Linton, Turkekul,
  Bejnood, Carver, Gopalan, Abida, Laudone, Biton, Chaudhary, Xu, Masilionis,
  Manova, Mazutis, Pe'er, Regev, and Sawyers]{karthaus2020regenerative}
Wouter~R Karthaus, Matan Hofree, Danielle Choi, Eliot~L Linton, Mesruh
  Turkekul, Alborz Bejnood, Brett Carver, Anuradha Gopalan, Wassim Abida,
  Vincent Laudone, Moshe Biton, Ojasvi Chaudhary, Tianhao Xu, Ignas Masilionis,
  Katia Manova, Linas Mazutis, Dana Pe'er, Aviv Regev, and Charles Sawyers.
\newblock Regenerative potential of prostate luminal cells revealed by
  single-cell analysis.
\newblock \emph{Science}, 368\penalty0 (6490):\penalty0 497--505, 2020.

\bibitem[Baccin et~al.(2019)Baccin, Al-Sabah, Velten, Helbling,
  Gr{\"u}nschl{\"a}ger, Hern{\'a}ndez-Malmierca, Nombela-Arrieta, Steinmetz,
  Trumpp, and Haas]{baccin2019combined}
Chiara Baccin, Jude Al-Sabah, Lars Velten, Patrick~M Helbling, Florian
  Gr{\"u}nschl{\"a}ger, Pablo Hern{\'a}ndez-Malmierca, C{\'e}sar
  Nombela-Arrieta, Lars~M Steinmetz, Andreas Trumpp, and Simon Haas.
\newblock Combined single-cell and spatial transcriptomics reveal the
  molecular, cellular and spatial bone marrow niche organization.
\newblock \emph{Nature Cell Biology}, pages 1--11, 2019.

\bibitem[Chen and Salman(2011)]{chen2011extracting}
Ke~Chen and Ahmad Salman.
\newblock Extracting speaker-specific information with a regularized siamese
  deep network.
\newblock In \emph{Advances in Neural Information Processing Systems}, pages
  298--306, 2011.

\bibitem[{Duque} et~al.(2019){Duque}, {Wolf}, and {Moon}]{duque2019}
Andr{\'e}s~F. {Duque}, Guy {Wolf}, and Kevin~R. {Moon}.
\newblock Visualizing high dimensional dynamical processes.
\newblock In \emph{IEEE International Workshop on Machine Learning for Signal
  Processing}, 2019.

\bibitem[Horoi et~al.(2020)Horoi, Geadah, Wolf, and Lajoie]{horoi2020low}
Stefan Horoi, Victor Geadah, Guy Wolf, and Guillaume Lajoie.
\newblock Low-dimensional dynamics of encoding and learning in recurrent neural
  networks.
\newblock In \emph{Canadian Conference on Artificial Intelligence}, pages
  276--282. Springer, 2020.

\bibitem[Gigante et~al.(2019)Gigante, Charles, Krishnaswamy, and
  Mishne]{gigante2019visualizing}
Scott Gigante, Adam~S Charles, Smita Krishnaswamy, and Gal Mishne.
\newblock Visualizing the phate of neural networks.
\newblock In \emph{Advances in Neural Information Processing Systems}, pages
  1840--1851, 2019.

\bibitem[Bair et~al.(2006)Bair, Hastie, Paul, and Tibshirani]{bair2006}
Eric Bair, Trevor Hastie, Debashis Paul, and Robert Tibshirani.
\newblock Prediction by supervised principal components.
\newblock \emph{Journal of the American Statistical Association}, 101\penalty0
  (473):\penalty0 119--137, 2006.

\bibitem[Yu et~al.(2006)Yu, Yu, Tresp, Kriegel, and Wu]{yu2006}
Shipeng Yu, Kai Yu, Volker Tresp, Hans-Peter Kriegel, and Mingrui Wu.
\newblock Supervised probabilistic principal component analysis.
\newblock In \emph{Proceedings of the 12th ACM SIGKDD International Conference
  on Knowledge Discovery and Data Mining}, KDD ’06, page 464–473, New York,
  NY, USA, 2006. Association for Computing Machinery.
\newblock ISBN 1595933395.

\bibitem[qing Zhang(2009)]{ZHANG2009}
Shi qing Zhang.
\newblock Enhanced supervised locally linear embedding.
\newblock \emph{Pattern Recognition Letters}, 30\penalty0 (13):\penalty0 1208
  -- 1218, 2009.
\newblock ISSN 0167-8655.

\bibitem[{Li} et~al.(2018){Li}, {Tang}, and {He}]{Li2018}
Z.~{Li}, J.~{Tang}, and X.~{He}.
\newblock Robust structured nonnegative matrix factorization for image
  representation.
\newblock \emph{IEEE Transactions on Neural Networks and Learning Systems},
  29\penalty0 (5):\penalty0 1947--1960, 2018.

\bibitem[{Zhang} et~al.(2016){Zhang}, {Zong}, {Liu}, and {Luo}]{Zhang2016}
X.~{Zhang}, L.~{Zong}, X.~{Liu}, and J.~{Luo}.
\newblock Constrained clustering with nonnegative matrix factorization.
\newblock \emph{IEEE Transactions on Neural Networks and Learning Systems},
  27\penalty0 (7):\penalty0 1514--1526, 2016.

\bibitem[{Jia} et~al.(2019){Jia}, {Kwong}, {Hou}, and {Wu}]{Jia2019}
Y.~{Jia}, S.~{Kwong}, J.~{Hou}, and W.~{Wu}.
\newblock Semi-supervised non-negative matrix factorization with dissimilarity
  and similarity regularization.
\newblock \emph{IEEE Transactions on Neural Networks and Learning Systems},
  pages 1--12, 2019.

\bibitem[Vlachos et~al.(2002)Vlachos, Domeniconi, Gunopulos, Kollios, and
  Koudas]{vlachos2002}
Michail Vlachos, Carlotta Domeniconi, Dimitrios Gunopulos, George Kollios, and
  Nick Koudas.
\newblock Non-linear dimensionality reduction techniques for classification and
  visualization.
\newblock In \emph{Proceedings of the Eighth ACM SIGKDD International
  Conference on Knowledge Discovery and Data Mining}, KDD ’02, page
  645–651, New York, NY, USA, 2002. Association for Computing Machinery.
\newblock ISBN 158113567X.

\bibitem[Ribeiro et~al.(2008)Ribeiro, Vieira, and Carvalho~das
  Neves]{Ribeiro2008}
Bernardete Ribeiro, Armando Vieira, and Jo{\~a}o Carvalho~das Neves.
\newblock Supervised isomap with dissimilarity measures in embedding learning.
\newblock In Jos{\'e} Ruiz-Shulcloper and Walter~G. Kropatsch, editors,
  \emph{Progress in Pattern Recognition, Image Analysis and Applications},
  pages 389--396, Berlin, Heidelberg, 2008. Springer Berlin Heidelberg.
\newblock ISBN 978-3-540-85920-8.

\bibitem[Tapson and Greene(2005)]{tapson2005}
J.~Tapson and J.R. Greene.
\newblock Plant data visualization using non-negative matrix factorization.
\newblock \emph{IFAC Proceedings Volumes}, 38\penalty0 (1):\penalty0 73 -- 78,
  2005.
\newblock ISSN 1474-6670.
\newblock 16th IFAC World Congress.

\bibitem[Babaee et~al.(2016)Babaee, Tsoukalas, Rigoll, and Datcu]{Babaee2016}
Mohammadreza Babaee, Stefanos Tsoukalas, Gerhard Rigoll, and Mihai Datcu.
\newblock Immersive visualization of visual data using nonnegative matrix
  factorization.
\newblock \emph{Neurocomputing}, 173:\penalty0 245 -- 255, 2016.
\newblock ISSN 0925-2312.

\bibitem[Barshan et~al.(2011)Barshan, Ghodsi, Azimifar, and
  Zolghadri~Jahromi]{barshan2011}
Elnaz Barshan, Ali Ghodsi, Zohreh Azimifar, and Mansoor Zolghadri~Jahromi.
\newblock Supervised principal component analysis: Visualization,
  classification and regression on subspaces and submanifolds.
\newblock \emph{Pattern Recogn.}, 44\penalty0 (7):\penalty0 1357–1371, July
  2011.
\newblock ISSN 0031-3203.

\bibitem[Breiman(2001)]{rf}
Leo Breiman.
\newblock Random forests.
\newblock \emph{Mach. Learn.}, 45\penalty0 (1):\penalty0 5–32, October 2001.
\newblock ISSN 0885-6125.

\bibitem[Chao et~al.(2019)Chao, Luo, and Ding]{Chao2019}
Guoqing Chao, Yuan Luo, and Weiping Ding.
\newblock Recent advances in supervised dimension reduction: A survey.
\newblock \emph{Machine Learning and Knowledge Extraction}, 1\penalty0
  (1):\penalty0 341–358, Jan 2019.
\newblock ISSN 2504-4990.

\bibitem[{Tsuge} et~al.(2001){Tsuge}, {Shishibori}, {Kuroiwa}, and
  {Kita}]{Tsuge2001}
S.~{Tsuge}, M.~{Shishibori}, S.~{Kuroiwa}, and K.~{Kita}.
\newblock Dimensionality reduction using non-negative matrix factorization for
  information retrieval.
\newblock In \emph{2001 IEEE International Conference on Systems, Man and
  Cybernetics. e-Systems and e-Man for Cybernetics in Cyberspace
  (Cat.No.01CH37236)}, volume~2, pages 960--965 vol.2, 2001.

\bibitem[{Haifeng Liu} et~al.(2012){Haifeng Liu}, {Zhaohui Wu}, {Xuelong Li},
  {Deng Cai}, and {Huang}]{Liu2012}
{Haifeng Liu}, {Zhaohui Wu}, {Xuelong Li}, {Deng Cai}, and T.~S. {Huang}.
\newblock Constrained nonnegative matrix factorization for image
  representation.
\newblock \emph{IEEE Transactions on Pattern Analysis and Machine
  Intelligence}, 34\penalty0 (7):\penalty0 1299--1311, 2012.

\bibitem[Liu et~al.(2008)Liu, Yuan, and Ye]{Liu2008}
Weixiang Liu, Kehong Yuan, and Datian Ye.
\newblock Reducing microarray data via nonnegative matrix factorization for
  visualization and clustering analysis.
\newblock \emph{Journal of Biomedical Informatics}, 41\penalty0 (4):\penalty0
  602 -- 606, 2008.
\newblock ISSN 1532-0464.

\bibitem[{Cai} et~al.(2008){Cai}, {He}, {Wu}, and {Han}]{Cai2008}
D.~{Cai}, X.~{He}, X.~{Wu}, and J.~{Han}.
\newblock Non-negative matrix factorization on manifold.
\newblock In \emph{2008 Eighth IEEE International Conference on Data Mining},
  pages 63--72, 2008.

\bibitem[van~der Maaten and Hinton(2008)]{vanDerMaaten2008}
Laurens van~der Maaten and Geoffrey Hinton.
\newblock Visualizing data using {t-SNE}.
\newblock \emph{Journal of Machine Learning Research}, 9:\penalty0 2579--2605,
  2008.

\bibitem[{Belkin} and {Niyogi}(2003)]{belkin2003}
M.~{Belkin} and P.~{Niyogi}.
\newblock Laplacian eigenmaps for dimensionality reduction and data
  representation.
\newblock \emph{Neural Computation}, 15\penalty0 (6):\penalty0 1373--1396,
  2003.

\bibitem[de~Ridder et~al.(2003)de~Ridder, Kouropteva, Okun, Pietik{\"a}inen,
  and Duin]{ridder2003}
Dick de~Ridder, Olga Kouropteva, Oleg Okun, Matti Pietik{\"a}inen, and Robert
  P.~W. Duin.
\newblock Supervised locally linear embedding.
\newblock In Okyay Kaynak, Ethem Alpaydin, Erkki Oja, and Lei Xu, editors,
  \emph{Artificial Neural Networks and Neural Information Processing ---
  ICANN/ICONIP 2003}, pages 333--341, Berlin, Heidelberg, 2003. Springer Berlin
  Heidelberg.
\newblock ISBN 978-3-540-44989-8.

\bibitem[Polito and Perona(2002)]{polito2002}
Marzia Polito and Pietro Perona.
\newblock Grouping and dimensionality reduction by locally linear embedding.
\newblock In T.~G. Dietterich, S.~Becker, and Z.~Ghahramani, editors,
  \emph{Advances in Neural Information Processing Systems 14}, pages
  1255--1262. MIT Press, 2002.

\bibitem[Friedman et~al.(1977)Friedman, Bentley, and Finkel]{friedman1977kd}
Jerome~H. Friedman, Jon~Louis Bentley, and Raphael~Ari Finkel.
\newblock An algorithm for finding best matches in logarithmic expected time.
\newblock \emph{ACM Trans. Math. Softw.}, 3\penalty0 (3):\penalty0 209–226,
  September 1977.
\newblock ISSN 0098-3500.

\bibitem[Cutler et~al.(2012)Cutler, Cutler, and Stevens]{Cutler2012}
Adele Cutler, D.~Richard Cutler, and John~R. Stevens.
\newblock \emph{Random Forests}, pages 157--175.
\newblock Springer US, Boston, MA, 2012.
\newblock ISBN 978-1-4419-9326-7.

\bibitem[Breiman(1996)]{Breiman1996}
Leo Breiman.
\newblock Bagging predictors.
\newblock \emph{Machine Learning}, 24\penalty0 (2):\penalty0 123--140, Aug
  1996.
\newblock ISSN 1573-0565.

\bibitem[Segal and Xiao(2011)]{segal2011}
Mark Segal and Yuanyuan Xiao.
\newblock Multivariate random forests.
\newblock \emph{WIREs Data Mining and Knowledge Discovery}, 1\penalty0
  (1):\penalty0 80--87, 2011.

\bibitem[Breiman and Cutler((Accessed on 04/15/2020))]{rfsd:Online}
Leo Breiman and Adele Cutler.
\newblock Random forests for scientific discovery.
\newblock \url{http://www.math.usu.edu/adele/RandomForests/ENAR.pdf}, (Accessed
  on 04/15/2020).

\bibitem[Lin and Jeon(2006)]{rfadaptiveknn}
Yi~Lin and Yongho Jeon.
\newblock Random forests and adaptive nearest neighbors.
\newblock \emph{Journal of the American Statistical Association}, 101\penalty0
  (474):\penalty0 578--590, 2006.

\bibitem[Golino and Gomes(2014)]{golino2014}
Hudson~F. Golino and Cristiano Mauro~Assis Gomes.
\newblock {Visualizing Random Forest's Prediction Results}.
\newblock \emph{Psychology}, 5:\penalty0 2084--2098, 2014.

\bibitem[Fruchterman and Reingold(1991)]{frucht1991}
Thomas M.~J. Fruchterman and Edward~M. Reingold.
\newblock Graph drawing by force-directed placement.
\newblock \emph{Software: Practice and Experience}, 21\penalty0 (11):\penalty0
  1129--1164, 1991.

\bibitem[Gajer et~al.(2004)Gajer, Goodrich, and Kobourov]{GAJER20043}
Pawel Gajer, Michael~T. Goodrich, and Stephen~G. Kobourov.
\newblock A multi-dimensional approach to force-directed layouts of large
  graphs.
\newblock \emph{Computational Geometry}, 29\penalty0 (1):\penalty0 3 -- 18,
  2004.
\newblock ISSN 0925-7721.
\newblock Special Issue on the 10th Fall Workshop on Computational Geometry,
  SUNY at Stony Brook.

\bibitem[Duque et~al.(2019)Duque, Wolf, and Moon]{duque2019visualizing}
Andr{\'e}s~F Duque, Guy Wolf, and Kevin~R Moon.
\newblock Visualizing high dimensional dynamical processes.
\newblock In \emph{2019 IEEE 29th International Workshop on Machine Learning
  for Signal Processing (MLSP)}, pages 1--6. IEEE, 2019.

\bibitem[Haghverdi et~al.(2016)Haghverdi, Buettner, Wolf, Buettner, and
  Theis]{haghverdi2016diffusion}
Laleh Haghverdi, Maren Buettner, F~Alexander Wolf, Florian Buettner, and
  Fabian~J Theis.
\newblock Diffusion pseudotime robustly reconstructs lineage branching.
\newblock \emph{Nature Methods}, 13\penalty0 (10):\penalty0 845, 2016.

\bibitem[Dua and Graff(2017)]{UCI2019}
Dheeru Dua and Casey Graff.
\newblock {UCI} machine learning repository, 2017.

\bibitem[Liaw and Wiener(2002)]{randomForest}
Andy Liaw and Matthew Wiener.
\newblock Classification and regression by randomforest.
\newblock \emph{R News}, 2\penalty0 (3):\penalty0 18--22, 2002.

\bibitem[{R Core Team}(2019)]{R}
{R Core Team}.
\newblock \emph{R: A Language and Environment for Statistical Computing}.
\newblock R Foundation for Statistical Computing, Vienna, Austria, 2019.

\bibitem[Xu et~al.(1992)Xu, Krzyzak, and Suen]{xu1992digits}
Lei Xu, A.~Krzyzak, and Ching Suen.
\newblock Methods of combining multiple classifiers and their applications to
  handwriting recognition.
\newblock \emph{IEEE Transactions on Systems Man and Cybernetics Part B
  (Cybernetics)}, 22:\penalty0 418 -- 435, 06 1992.

\bibitem[Cock(2011)]{cock2011ames}
Dean~De Cock.
\newblock Ames, iowa: Alternative to the boston housing data as an end of
  semester regression project.
\newblock \emph{Journal of Statistics Education}, 19\penalty0 (3):\penalty0
  null, 2011.

\bibitem[Wold et~al.(1984)Wold, Ruhe, Wold, and Dunn]{wold1984}
S.~Wold, A.~Ruhe, H.~Wold, and W.~J. Dunn, III.
\newblock The collinearity problem in linear regression. the partial least
  squares (pls) approach to generalized inverses.
\newblock \emph{SIAM Journal on Scientific and Statistical Computing},
  5\penalty0 (3):\penalty0 735--743, 1984.

\bibitem[Anderson(1936)]{anderson1936iris}
Edgar Anderson.
\newblock The species problem in iris.
\newblock \emph{Annals of the Missouri Botanical Garden}, 23\penalty0
  (3):\penalty0 457--509, 1936.
\newblock ISSN 00266493.

\bibitem[Gorman and Sejnowski(1988)]{GORMAN198875sonar}
R.Paul Gorman and Terrence~J. Sejnowski.
\newblock Analysis of hidden units in a layered network trained to classify
  sonar targets.
\newblock \emph{Neural Networks}, 1\penalty0 (1):\penalty0 75 -- 89, 1988.
\newblock ISSN 0893-6080.

\bibitem[Kuhn(2020)]{kuhn2020caret}
Max Kuhn.
\newblock \emph{caret: Classification and Regression Training}, 2020.
\newblock R package version 6.0-86.

\bibitem[{Moon} et~al.(2017){Moon}, {Sricharan}, and {Hero}]{moon2017ensMI}
K.~R. {Moon}, K.~{Sricharan}, and A.~O. {Hero}.
\newblock Ensemble estimation of mutual information.
\newblock In \emph{2017 IEEE International Symposium on Information Theory
  (ISIT)}, pages 3030--3034, 2017.

\bibitem[Tibshirani(1994)]{tibshirani1994lasso}
Robert Tibshirani.
\newblock Regression shrinkage and selection via the lasso.
\newblock \emph{JOURNAL OF THE ROYAL STATISTICAL SOCIETY, SERIES B},
  58:\penalty0 267--288, 1994.

\bibitem[D{\'i}az-Uriarte and Alvarez~de Andr{\'e}s(2006)]{uriarte2006}
Ram{\'o}n D{\'i}az-Uriarte and Sara Alvarez~de Andr{\'e}s.
\newblock Gene selection and classification of microarray data using random
  forest.
\newblock \emph{BMC Bioinformatics}, 7\penalty0 (1):\penalty0 3, Jan 2006.
\newblock ISSN 1471-2105.

\end{thebibliography}

\appendix
\section*{Appendices}

In these appendices, we provide further details about RF-PHATE. In Section~\ref{section:data}, we provide details about each of the data sets used in the experiments. We then provide details on the metric we use to quantify our results and show extended results in Section~\ref{section:varregress}. In Section~\ref{section:importance}, we present alternate measures of variable importance which are consistent with the random forest measure of variable importance. We compare RF-PHATE to unsupervised visualization methods in Section~\ref{sec:unsupervised}. We then perform an ablation study in Section~\ref{section:paramselection}, demonstrating that RF-PHATE is robust to the choice of parameters. Finally, we demonstrate that other approaches to embedding the random forest proximities result in inferior visualizations in Section~\ref{sec:prox}.

\section{Description of Data Sets}\label{section:data}

Descriptions of each the data sets used in the experiments and their preprocessing steps are given here.  In addition to steps listed particular to each data set, all continuous variables in each data set were standardized to have a mean of zero and a variance of one.

The \textit{Optical Digits}~\cite{xu1992digits, UCI2019} data contains 5620 observations of handwritten numerical digits (0 - 9) recorded in $32\times32$ bit images. The images were collect from 43 individuals and the pixels are summed into $4\times4$ blocks resulting in 64 dimensional arrays.  Each variable takes on integer values from 0 to 16 which denote the number of original pixels with writing included in that particular block.  There are no missing values.

The \textit{Ames Housing}~\cite{cock2011ames} data set consists of 1460 house listings with 80 variable with a mixture of nominal, ordinal, discrete, and continuous values. The response variable is the sales price. Missing values were manually imputed (N/A values were replaced with ``none'' for missing categorical variables, or 0 for continuous variables, where applicable). One-hot encoding was applied to categorical variables for compatibility with the compared dimensionality reduction techniques.

The \textit{Titanic} data set has 12 variables and 891 observations.  The variables Ticket and Name are unique to each observations and were thus removed.  In addition, observations with missing values were also removed, resulting in a total of 712 observations used in the comparisons.  One-hot encoding was applied to the categorical variables.

\begin{figure}[!htb]
    \centering
    \includegraphics[width = 0.75\textwidth]{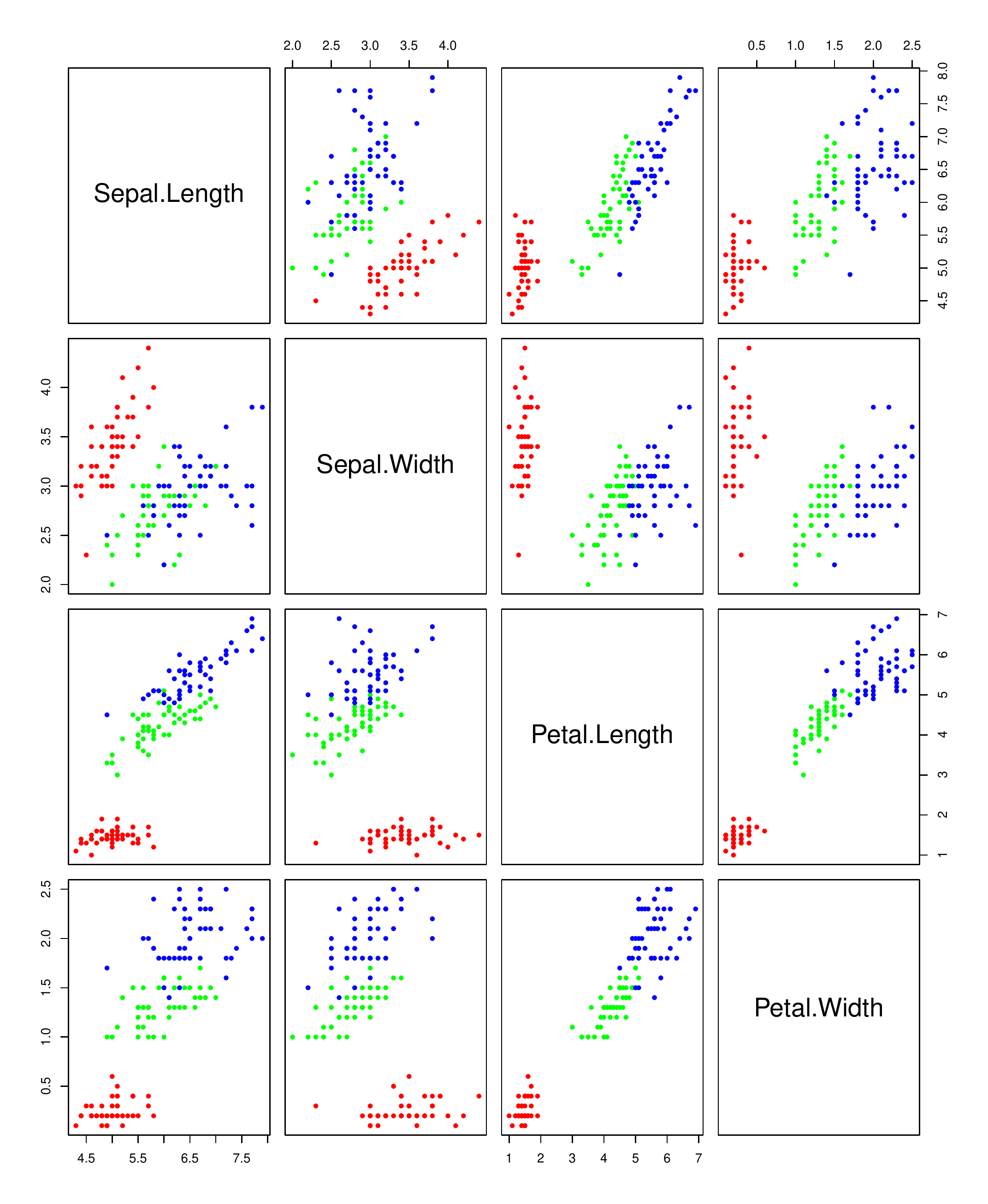}
    \caption{Pairwise variable plot of \textit{Iris}}
    \label{fig:irisPairs}
\end{figure}

\textit{Iris}~\cite{anderson1936iris, UCI2019} contains 150 observations of three iris species: virginica, versicolor, and setosa.  There are 50 observations from each species. Each observation has four measurements: sepal length, sepal width, petal length, and petal width.  The data set has no missing values. See a pair-wise variable plot in figure \ref{fig:irisPairs}.

\begin{figure}[!htb]
    \centering
    \includegraphics[width = 0.9\textwidth]{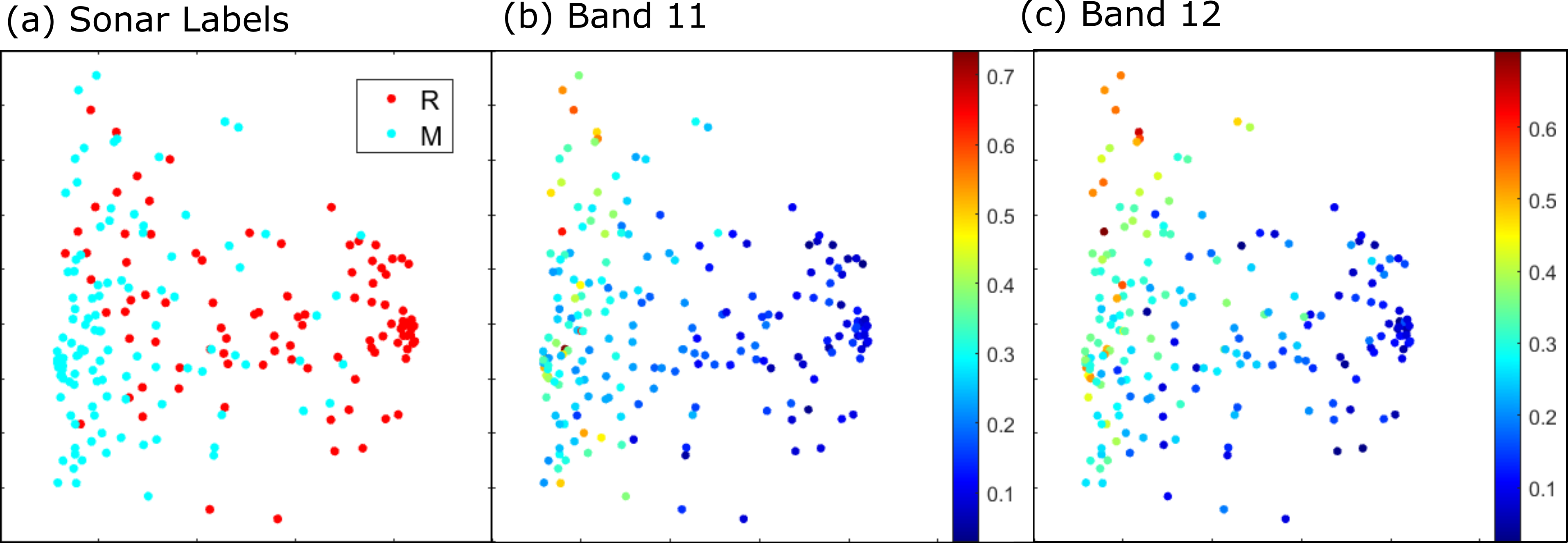}
    \caption{RF-PHATE embeddings of the \textit{Sonar} data set showing (a) class labels (M for metal, R for rock), (b) the top important variable (Band 11), and (c) the second important variable (Band 12). More densely clustered metal points on the left (blue) and rock points on the right (red) give an indication of random forest confidence in the predictions.  Sparsely scattered points in the center are more difficult to classify.}
    \label{fig:sonar}
\end{figure}

In addition to the previous data sets used in the main paper, we include results on other data sets. This includes the \textit{Sonar}~\cite{GORMAN198875sonar} data set, which consists of 208 instances of sonar signals reflected off of rocks or metal cylinders positioned at various angles. This results in 60 variables (60 sonar bands). Classification is done on the material type (rock or metal). The data set contains no missing values. (See figure \ref{fig:sonar}).

\begin{figure}[!htb]
    \centering
    \includegraphics[width = 0.9\textwidth]{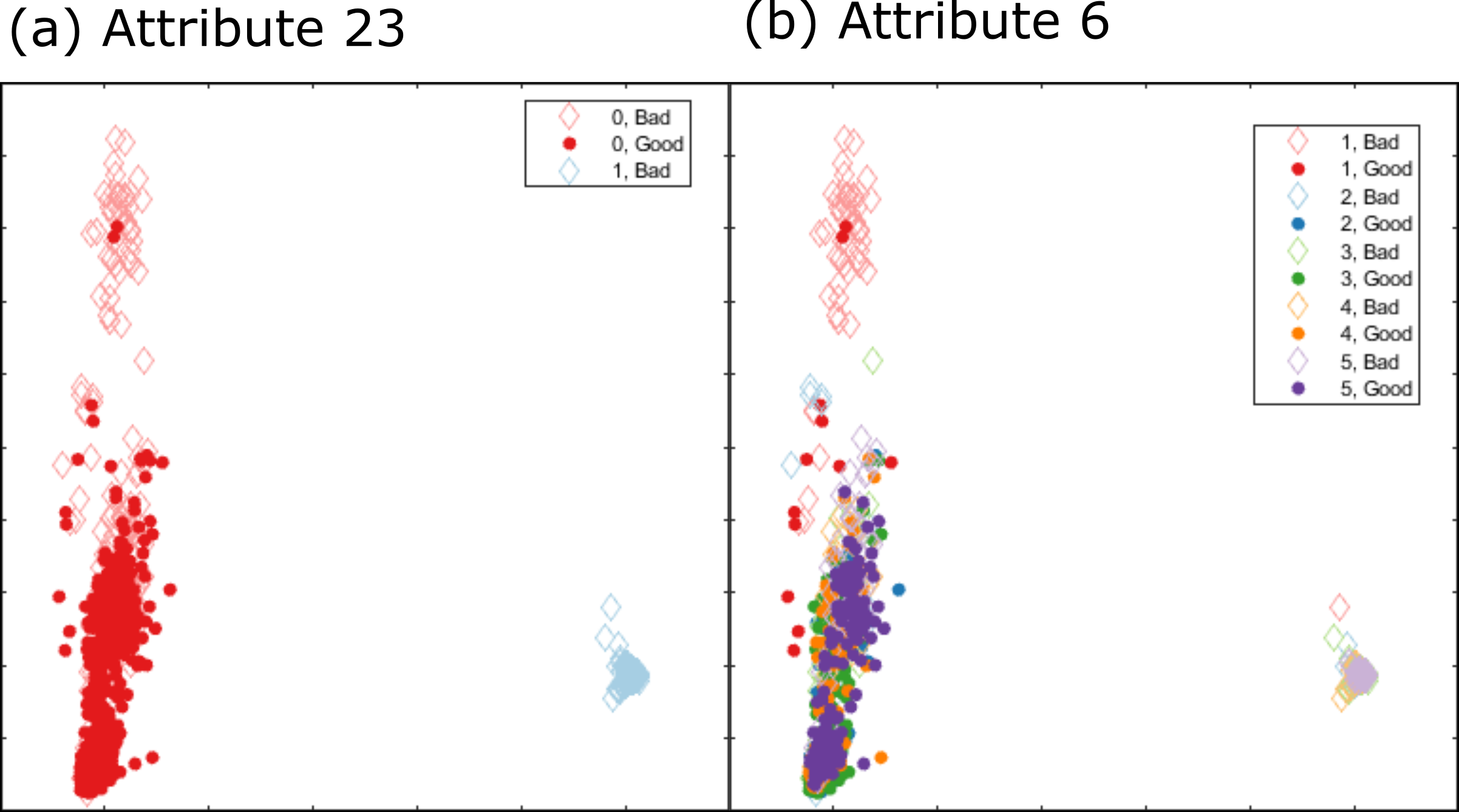}
    \caption{RF-PHATE embeddings of the \textit{German} data set. In (a) it is clear that a result of 1 in attribute 23 guarantees bad credit.  In (b) group 1 of attribute 6 further separates the majority of the remaining bad credit customers from those with good credit. Attribute values are denoted by different colors.}
    \label{fig:german}
\end{figure}

Additionally, we show results for the \textit{German Credit} data set~\cite{UCI2019}, consisting of 1000 instances of 20 categorical and numeric data sets.  An alternative version (that we use) contains 24 numeric variables representative of the original data.  The variables (attributes) are proprietary and therefore variable names are not included. The data set has two classes, bad credit (0) or good credit (1). RF-PHATE demonstrates variables which may be used to clearly distinguish customers with good credit from those with bad.  See figure~\ref{fig:german}.

\section{Low-Dimensional Variable Regression}\label{section:varregress}

To quantify the degree in which the low-dimensional embeddings captured important variables, we used the embeddings as feature matrices for a $k$-NN classifier/regressor on the original variables (see Tables 1 and 2). The value of $k$ used in each case was the square-root of the number of observations in the data set. For discrete variables, the average error rate was computed using 10-fold cross validation.  For continuous variables, the average root mean squared error (RMSE) was the criterion, also using 10-fold cross validation. We compared results using 7 supervised dimensionality reduction techniques, RF-PHATE, Enhanced Supervised Isomap~\cite{Ribeiro2008}, Kernalized Supervised PCA~\cite{yu2006}, Supervised PCA~\cite{yu2006}, Enhanced Supervised LLE~\cite{ZHANG2009}, Supervised NMF~\cite{Jia2019}, and Partial Least Squares Discriminant Analysis (PLSDA)~\cite{wold1984}.

Here we show extended results from those in the paper by including other datasets and by assessing the embeddings with dimensions of 4 and 5. The results are contained in Tables~\ref{tab:ames} to \ref{tab:german}. Here we see that RF-PHATE again outperforms the other methods, as it is mostly first and either second or third for all comparisons. No other method is as consistent in its performance across all datasets.

\begin{table}[!htb]
\centering
\caption{Variable regression on \textit{Ames Housing} data set.  ESIso, ESLLE, and SNMF have not been adapted for continuous labels, hence the missing values (N/A). \label{tab:ames}}
\begin{tabular}{|c|c|c|c|c|c|c|c|}
    \hline
    Var (dims)	&	RF-PHATE		&	ESIso		&	KSPCA		&	SPCA		&	ESLLE		&	SNMF		&	PLSDA		\\
    \hline
    Quality (2)	&	\underline{\textbf{0.645}}	&	N/A	&	0.815	&	0.765	&	N/A	&	N/A	&	\textbf{0.767}	\\
    Quality (3)	&	\underline{\textbf{0.631}}	&	N/A	&	0.789	&	0.760	&	N/A	&	N/A	&	\textbf{0.736}	\\
    Quality (4)	&	\underline{\textbf{0.619}}	&	N/A	&	0.785	&	0.747	&	N/A	&	N/A	&	\textbf{0.723}	\\
    Quality (5)	&	\underline{\textbf{0.619}}	&	N/A	&	0.783	&	0.743	&	N/A	&	N/A	&	\textbf{0.728}	\\
    \hline															
    GrLivArea (2)	&	\underline{\textbf{263.751}}	&	N/A	&	\textbf{402.084}	&	413.366	&	N/A	&	N/A	&	415.903	\\
    GrLivArea (3)	&	\underline{\textbf{224.579}}	&	N/A &	393.641	&	411.534	&	N/A	&	N/A	&	\textbf{293.184}	\\
    GrLivArea (4)	&	\underline{\textbf{226.168}}	&	N/A	&	390.608	&	385.940	&	N/A	&	N/A	&	\textbf{293.578}	\\
    GrLivArea (5)	&	\underline{\textbf{226.585}}	&	N/A	&	392.209	&	385.254	&	N/A	&	N/A &	\textbf{297.176}	\\
    \hline															
    Ext. Quality (2)	&	\underline{\textbf{0.114}}	&	N/A	&	\textbf{0.132}	&	0.134 & N/A	&	N/A	&	\textbf{0.125}	\\
    Ext. Quality (3)	&	\underline{\textbf{0.114}}	&	N/A	&	0.136	&	0.127	&	N/A	&	N/A	&	\textbf{0.125}	\\

    Ext. Quality (4)	&	\textbf{0.115}	&	N/A	&	0.124	&	0.128	&	N/A	&	N/A	&	\underline{\textbf{0.112}}	\\
    Ext. Quality (5)	&	\underline{\textbf{0.112}}	&	N/A	&	\textbf{0.123}	&	0.129	&	N/A	&	N/A	&	\underline{\textbf{0.112}}	\\
    
    \hline
\end{tabular}
\end{table}

\begin{table}[!htb]
\centering
\caption{Variable regression on \textit{Titanic}. Sex and Class are both categorical; the average 10-fold cross-validation error is reported.}
\begin{tabular}{|c|c|c|c|c|c|c|c|}
    \hline
    Var (dims)	&	RF-PHATE		&	ESIso		&	KSPCA		&	SPCA		&	ESLLE		&	SNMF		&	PLSDA		\\
    \hline
    
    Sex (2) & \underline{\textbf{0.0000}}	&	\textbf{0.0028}	&	0.0084	&	0.0056	&	0.0070	&	0.1419	&	\textbf{0.0028}	\\
    Sex (3) & \underline{\textbf{0.0000}}	&	\textbf{0.0028}	&	0.0084	&	\textbf{0.0028}	&	0.0042	&	0.1489	&	0.0042	\\
    Sex (4) & \underline{\textbf{0.0000}}	&	0.0070	&	0.0056	&	\textbf{0.0028}	&	\textbf{0.0028}	&	0.1320	&	0.0042	\\
    Sex (5) & \underline{\textbf{0.0000}}	&	0.0056	&	0.0056	&	\textbf{0.0028}	&	\textbf{0.0028}	&	0.1629	&	\textbf{0.0028}	\\
    \hline
    Class (2) & \underline{\textbf{0.0154}}	&	0.1657	&	0.1025	&	0.0688	&	0.0702	&	0.3174	&	\textbf{0.0197}	\\
    Class (3) & \underline{\textbf{0.0112}}	&	0.0492	&	0.1011	&	0.0281	&	0.0969	&	0.3048	&	\textbf{0.0239}	\\
    Class (4) & \underline{\textbf{0.0169}}	&	0.0421	&	0.0421	&	0.0253	&	0.0702	&	0.3666	&	\textbf{0.0211}	\\
    Class (5) & \underline{\textbf{0.0126}}	&	0.0463	&	\textbf{0.0197}	&	0.0253	&	0.0337	&	0.2893	&	\textbf{0.0197}	\\
        
\hline
\end{tabular}
\end{table}

\begin{table}[!htb]
\centering
\caption{Variable regression on \textit{Optical Digits}. The variables correspond to block locations in the condensed ($8\times 8)$ images. \label{tab:digits}}
\begin{tabular}{|c|c|c|c|c|c|c|c|}
    \hline
    Var (dims)	&	RF-PHATE		&	ESIso		&	KSPCA		&	SPCA		&	ESLLE		&	SNMF		&	PLSDA		\\
    \hline
    V43 (2)	&	\underline{\textbf{3.612}}	&	\textbf{3.554}	&	5.478	&	5.237	&	5.910	&	6.659	&	5.842	\\
    V43 (3)	&	\underline{\textbf{3.473}}	&	\textbf{3.484}	&	4.013	&	4.922	&	5.887	&	6.659	&	5.407	\\
    V43 (4)	&	\underline{\textbf{3.221}}	&	3.365	&	\textbf{3.266}	&	4.153	&	5.729	&	6.659	&	4.704	\\
    V43 (5)	&	\textbf{3.313}	&	3.386	&	\underline{\textbf{2.895}}	&	3.970	&	5.750	&	6.659	&	4.120	\\
    \hline															
    V22 (2)	&	\underline{\textbf{3.425}}	&	\textbf{3.869}	&	4.101	&	4.862	&	5.441	&	6.540	&	5.730	\\
    V22 (3)	&	\underline{\textbf{3.361}}	&	\textbf{3.851}	&	3.902	&	4.007	&	5.194	&	6.540	&	5.060	\\
    V22 (4)	&	\textbf{3.381}	&	3.765	&	\underline{\textbf{3.258}}	&	3.750	&	5.049	&	6.540	&	4.832	\\
    V22 (5)	&	\textbf{3.321}	&	3.712	&	\underline{\textbf{3.156}}	&	3.754	&	5.077	&	6.540	&	4.478	\\
    \hline															
    V44 (2)	&	\textbf{4.147}	&	\underline{\textbf{3.997}}	&	5.107	&	4.416	&	5.839	&	7.147	&	6.217	\\
    V44 (3)	&	\underline{\textbf{3.766}}	&	\textbf{3.810}	&	4.608	&	4.039	&	5.810	&	7.147	&	5.968	\\
    V44 (4)	&	\textbf{3.549}	&	3.735	&	\underline{\textbf{3.082}}	&	3.681	&	5.501	&	7.147	&	5.183	\\
    V44 (5)	&	3.614	&	3.724	&	\underline{\textbf{3.015}}	&	\textbf{3.459}	&	5.579	&	7.147	&	4.756	\\

\hline
\end{tabular}
\end{table}

\begin{table}[!htb]
\centering
\caption{Variable regression on \textit{Sonar} using the top two important variables (see section \ref{section:importance}). \label{tab:sonar}}
\begin{tabular}{|c|c|c|c|c|c|c|c|}
    \hline
    Var (dims)	&	RF-PHATE		&	ESIso		&	KSPCA		&	SPCA		&	ESLLE		&	SNMF		&	PLSDA		\\
    \hline
    Band 11 (2) & \underline{\textbf{0.0889}}	&	0.1224	&	\textbf{0.0918}	&	0.0978	&	0.1173	&	0.1158	&	0.0958	\\
    Band 11 (3) & \underline{\textbf{0.0874}}	&	0.1217	&	\textbf{0.0907}	&	0.0985	&	0.1143	&	0.1209	&	0.0924	\\
    Band 11 (4) & \underline{\textbf{0.0873}}	&	0.1179	&	\textbf{0.0898}	&	0.0973	&	0.1157	&	0.1242	&	0.0905	\\
    Band 11 (5) & \underline{\textbf{0.0874}}	&	0.1185	&	\textbf{0.0877}	&	0.0972	&	0.1080	&	0.1232	&	0.0880	\\
    \hline
    Band 12 (2) & \underline{\textbf{0.0957}}	&	0.1327	&	0.1119	&	\textbf{0.1104}	&	0.1278	&	0.1195	&	0.1158	\\
    Band 12 (3) & \underline{\textbf{0.0949}}	&	0.1309	&	0.1112	&	\textbf{0.1097}	&	0.1229	&	0.1248	&	0.1139	\\
    Band 12 (4) & \underline{\textbf{0.0946}}	&	0.1304	&	0.1110	&	0.1089	&	0.1214	&	0.1308	&	\textbf{0.1053}	\\
    Band 12 (5) & \underline{\textbf{0.0951}}	&	0.1316	&	0.1103	&	0.1088	&	0.1102	&	0.1356	&	\textbf{0.0984}	\\
    \hline
\end{tabular}
\end{table}

\begin{table}[!htb]
\centering
\caption{Variable regression on the \textit{German} data set.\label{tab:german}}
\begin{tabular}{|c|c|c|c|c|c|c|c|}
    \hline
    Var (dims)	&	RF-PHATE		&	ESIso		&	KSPCA		&	SPCA		&	ESLLE		&	SNMF		&	PLSDA		\\
    \hline
    V23 (2)	&	\underline{\textbf{0.000}}	&	0.061	&	0.027	&	\textbf{0.025}	&	0.198	&	0.170	&	\textbf{0.025}	\\
    V23 (3)	&	\underline{\textbf{0.000}}	&	0.059	&	0.027	&	\textbf{0.024}	&	0.176	&	0.170	&	0.028	\\
    V23 (4)	&	\underline{\textbf{0.000}}	&	0.042	&	0.027	&	\textbf{0.024}	&	0.063	&	0.170	&	0.026	\\
    V23 (5)	&	\underline{\textbf{0.000}}	&	0.036	&	0.027	&	\textbf{0.025}	&	0.063	&	0.170	&	0.027	\\
    \hline															
    V6 (2)	&	\underline{\textbf{0.551}}	&	0.621	&	\textbf{0.596}	&	0.666	&	0.684	&	0.657	&	0.680	\\
    V6 (3)	&	\underline{\textbf{0.541}}	&	0.590	&	0.577	&	0.668	&	0.594	&	\textbf{0.560}	&	0.674	\\
    V6 (4)	&	\underline{\textbf{0.510}}	&	0.594	&	\underline{\textbf{0.510}}	&	0.646	&	\textbf{0.568}	&	0.611	&	0.637	\\
    V6 (5)	&	\underline{\textbf{0.510}}	&	0.567	&	\textbf{0.515}	&	0.643	&	0.579	&	0.518	&	0.606	\\
    \hline
\end{tabular}
\end{table}

\section{Alternative Assessments of Variable Importance}\label{section:importance}

To show that our measure of variable importance using random forests is robust, we computed variable importance  using several other methods: ROC curve variable importance (for discrete labels) or $R^2$ importance (for continuous labels) using the \texttt{caret}~\cite{kuhn2020caret} package in \texttt{R}~\cite{R} as well as using a mutual information (MI) estimator based on~\cite{moon2017ensMI}.  The results are given in table~\ref{table:varImp}. The \texttt{varImp} function from the \texttt{caret} package\cite{kuhn2020caret} used repeated 10-fold cross validation with 3 repeats. The method used was LVQ (learning vector quantization) for discrete responses, and LASSO~\cite{tibshirani1994lasso} ($R^2$ importance) for continuous responses. 

We note that the results are generally consistent with each other, especially when comparing the random forest results with the ROC or $R^2$ measure. While the MI gives some slight differences in some of the datasets, they are still generally the same. For example, MI determines the top three variables for the Titanic dataset to be Age, Fare, and Sex, respectively. The other two methods include Sex and Fare in their top three as well, albeit in slightly different orders. Overall, the results indicate that we are selecting important variables for the experiments.

\begin{table}[!htb]
    \centering
    \caption{Variable importance ranking as assessed by random forests, ROC curve importance (or $R^2$ importance for continuous responses), and mutual information estimation.  This mutual information estimation implementation is currently only written for discrete response variables. \label{table:varImp}}
    \begin{tabular}{|c||c|c|c|c|}
    \hline
    Data Set & Rank & RF & ROC / $R^2$ & MI \\
    \hline
    \hline
    \textit{Iris} & 1 & Petal Length & Petal Length & Petal Length\\
    \hline
    \textit{Iris} & 2 & Petal Width & Petal Width & Petal Width\\
    \hline
    \textit{Iris} & 3 & Sepal Length & Sepal Width & Sepal Length \\
    \hline
    \hline
    \textit{Optical Digits} & 1 & V43 & V22 & V43 \\
    \hline
    \textit{Optical Digits} & 2 & V22 & V43 & V22\\
    \hline
    \textit{Optical Digits} & 3 & V44 & V31 & V31\\
    \hline
    \hline
    \textit{Titanic} & 1 & Sex & Sex & Age\\
    \hline
    \textit{Titanic} & 2 & Class & Fare & Fare\\
    \hline
    \textit{Titanic} & 3 & Fare & Class & Sex\\
    \hline
    \hline
    \textit{Ames Housing} & 1 & Quality &  Quality & N/A\\
    \hline
    \textit{Ames Housing} & 2 & GrLivArea &  GrLivArea & N/A\\
    \hline
    \textit{Ames Housing} & 3 & Ext. Quality &  TotalBsmtSF & N/A\\
    \hline
    \hline
    \textit{Sonar} & 1 & Band 11 & Band 11 & Band 9\\
    \hline
    \textit{Sonar} & 2 & Band 12 & Band 12 & Band 11\\
    \hline
    \textit{Sonar} & 3 & Band 9 & Band 10 & Band 12\\
    \hline
    \hline
    \textit{German} & 1 & V23 & V23 & V4\\
    \hline
    \textit{German} & 2 & V6 & V10 & V10\\
    \hline
    \textit{German} & 3 & V22 & V6 & V23\\
    \hline
\end{tabular}
\end{table}

\section{Unsupervised Results on Noisy Data}
\label{sec:unsupervised}
Data exploration with excessive noise variables provides an example of a use-case scenario for supervised dimensionality reduction. In this section, we include regression results for the \textit{Iris} dataset with 1000 additional noise variables. The noise variables were randomly generated from Gaussian distributions with means uniformly sampled from values between -1 and 1. Each had a variance of 1. All variables (noise and otherwise) were scaled and centered prior to performing dimensionality reduction. The low-dimensional embeddings from 7 supervised and 7 unsupervised dimensionality techniques were used as predictors while the original data variables were used as the response for $k$-NN regression (unweighted). We used the square root of the number of observations as the $k$ parameter. The average RMSE using 10-fold cross validation was recorded for each regression problem with the experiment  repeated 10 times. The same ten data sets were used for both the supervised and unsupervised cases. 

Tables \ref{tab:irisnoisy} and \ref{tab:irisUnsup} show the supervised and unsupervised results, respectively. First, we see that RF-PHATE universally outperforms all of the supervised and unsupervised methods, indicating that RF-PHATE is very useful when there are many noise variables. We also observe that all of the unsupervised methods perform worse than all of the supervised methods except for SNMF. Thus supervised methods for visualization are preferred when the data contain noise variables.

\begin{table}[!htb]
\scriptsize
\centering
\caption{Variable regression on \textit{Iris} with 1000 additional random noise variables.  The low-dimensional embeddings of seven supervised dimensionality reduction methods were used as features to regress on the original iris data set variables. The noise variables were simulated ten times and  were averaged. The standard deviation is also reported. RF-PHATE outperforms all other methods. \label{tab:irisnoisy}}
\begin{tabular}{|c|c|c|c|c|c|c|c|}
\hline
    Var (dims)	&	RF-PHATE		&	ESIso		&	KSPCA		&	SPCA		&	ESLLE		&	SNMF		&	PLSDA		\\

    \hline
     petal len. (2)	&	\underline{\textbf{0.330}}	$\pm$ \underline{\textbf{0.01}}	&	0.348	$\pm$ 0.01	&	\textbf{0.340}	$\pm$ \textbf{0.01}	&	0.381	$\pm$ 0.01	&	0.349	$\pm$ 0.01	&	1.475	$\pm$ 0.03	&	0.387	$\pm$ 0.01	\\
    petal len. (3)	&	\underline{\textbf{0.334}}	$\pm$ \underline{\textbf{0.01}}	&	\textbf{0.353}	$\pm$ \textbf{0.01}	&	0.360	$\pm$ 0.01	&	0.400	$\pm$ 0.01	&	\textbf{0.353}	$\pm$ \textbf{0.01}	&	2.233	$\pm$ 0.04	&	0.363	$\pm$ 0.01	\\
    petal len. (4)	&	\underline{\textbf{0.337}} $\pm$ \underline{\textbf{0.01}}	&	0.355$\pm$0.01	&	0.361$\pm$0.01	&	0.412$\pm$0.01	&	\textbf{0.352} $\pm$ \textbf{0.01}	&	2.233$\pm$0.04	&	0.369$\pm$0.01	\\

    \hline	
    
    petal wid. (2)	&	\underline{\textbf{0.291}}	$\pm$ \underline{\textbf{0.01}}	&	0.298	$\pm$ 0.01	&	\textbf{0.297}	$\pm$ \textbf{0.01}	&	0.330	$\pm$ 0.01	&	0.304	$\pm$ 0.01	&	1.297	$\pm$ 0.02	&	0.330	$\pm$ 0.01	\\
    petal wid. (3)	&	\underline{\textbf{0.292}}	$\pm$ \underline{\textbf{0.01}}	&	\textbf{0.305}	$\pm$ \textbf{0.01}	&	0.312	$\pm$ 0.01	&	0.345	$\pm$ 0.01	&	0.308	$\pm$ 0.01	&	1.905	$\pm$ 0.03	&	0.308	$\pm$ 0.01	\\
    petal wid. (4)	&	\underline{\textbf{0.290}} $\pm$ \underline{\textbf{0.01}}	&	\textbf{0.304} $\pm$ \textbf{0.01}	&	0.314$\pm$0.01	&	0.358$\pm$0.01	&	0.306$\pm$0.01	&	1.905$\pm$0.03	&	0.321$\pm$0.01	\\

    \hline
    
    sepal len. (2)	&	\underline{\textbf{0.459}}	$\pm$ \underline{\textbf{0.01}}	&	0.522	$\pm$ 0.02	&	\textbf{0.514}	$\pm$ \textbf{0.02}	&	\textbf{0.514}	$\pm$ \textbf{0.02}	&	0.528	$\pm$ 0.02	&	0.885	$\pm$ 0.02	&	\textbf{0.514}	$\pm$ \textbf{0.02}	\\
    sepal len. (3)	&	\underline{\textbf{0.455}}	$\pm$ \underline{\textbf{0.01}}	&	0.536	$\pm$ 0.02	&	0.540	$\pm$ 0.02	&	0.520	$\pm$ 0.02	&	0.534	$\pm$ 0.02	&	1.272	$\pm$ 0.02	&	\textbf{0.516}	$\pm$ \textbf{0.02}	\\
    sepal len. (4)	&	\underline{\textbf{0.456}} $\pm$ \underline{\textbf{0.01}}	&	0.537$\pm$0.02	&	0.541$\pm$0.02	&	0.522$\pm$0.02	&	0.533$\pm$0.02	&	1.272$\pm$0.02	&	\textbf{0.517} $\pm$ \textbf{0.02}	\\

    \hline
    
    sepal wid. (2)	&	\underline{\textbf{0.320}}	$\pm$ \underline{\textbf{0.01}}	&	\textbf{0.348}	$\pm$ \textbf{0.01}	&	0.351	$\pm$ 0.01	&	0.357	$\pm$ 0.01	&	0.349	$\pm$ 0.01	&	0.462	$\pm$ 0.01	&	0.357	$\pm$ 0.01	\\
    sepal wid. (3)	&	\underline{\textbf{0.318}}	$\pm$ \underline{\textbf{0.01}}	&	\textbf{0.354}	$\pm$ \textbf{0.01}	&	0.358	$\pm$ 0.01	&	0.356	$\pm$ 0.01	&	\textbf{0.354}	$\pm$ \textbf{0.01}	&	0.501	$\pm$ 0.02	&	0.357	$\pm$ 0.01	\\
    sepal wid. (4)	&	\underline{\textbf{0.321}} $\pm$ \underline{\textbf{0.01}}	&	0.355$\pm$0.01	&	0.361$\pm$0.01	&	0.357$\pm$0.01	&	\textbf{0.354} $\pm$ \textbf{0.01}	&	0.501$\pm$0.02	&	0.358$\pm$0.01	\\

\hline
\end{tabular}
\end{table}

\begin{table}
\scriptsize
\centering
\caption{Variable regression on \textit{Iris} with 1000 additional random Gaussian noise variables.  The low-dimensional embeddings of seven unsupervised dimensionality reduction methods were used as features to regress on the original iris data set variables as a comparison to the supervised methods in table \ref{tab:irisnoisy}. The experiments were repeated 10 times. Compared to the supervised results, regression using the unsupervised embeddings performed much worse than all supervised cases with the exception of SNMF. \label{tab:irisUnsup}}
\begin{tabular}{|c|c|c|c|c|c|c|c|}
    \hline
    Dimensions	&	PHATE	&	ISO	&	PCA	&	LLE	&	NMF	& MDS &	TSNE\\
    \hline
    petal len. (2)	&	1.393 $\pm$ 0.03	&	1.398 $\pm$ 0.03	&	1.366 $\pm$ 0.03	&	1.365 $\pm$ 0.03	&	1.768 $\pm$ 0.03	&	\underline{\textbf{1.322}} $\pm$ \underline{\textbf{0.03}}	&	\textbf{1.355} $\pm$ \textbf{0.03}	\\
    petal len. (3)	&	1.372 $\pm$ 0.03	&	1.393 $\pm$ 0.03	&	1.366 $\pm$ 0.03	&	\underline{\textbf{1.339}} $\pm$ \underline{\textbf{0.03}}	&	1.539 $\pm$ 0.03	&	\textbf{1.342} $\pm$ \textbf{0.03}	&	1.355 $\pm$ 0.03	\\
    petal len. (4)	&	1.355 $\pm$ 0.03	&	1.365 $\pm$ 0.03	&	1.366 $\pm$ 0.03	&	1.34 $\pm$ 0.03	&	1.388 $\pm$ 0.03	&	\underline{\textbf{1.307}} $\pm$ \underline{\textbf{0.03}}	&	\textbf{1.328} $\pm$ \textbf{0.03}	\\
    \hline															

    petal wid. (2)	&	1.196 $\pm$ 0.02	&	1.181 $\pm$ 0.02	&	\textbf{1.120} $\pm$ \textbf{0.02}	&	1.159 $\pm$ 0.02	&	1.502 $\pm$ 0.03	&	\underline{\textbf{1.119}} $\pm$ \underline{\textbf{0.02}}	&	1.164 $\pm$ 0.02	\\
    petal wid. (3)	&	1.170 $\pm$ 0.02	&	1.183 $\pm$ 0.02	&	\textbf{1.120} $\pm$ \textbf{0.02}	&	1.157 $\pm$ 0.02	&	1.300 $\pm$ 0.02	&	\underline{\textbf{1.113}} $\pm$ \underline{\textbf{0.02}}	& 1.156 $\pm$ 0.02	\\
    petal wid. (4)	&	1.163 $\pm$ 0.02	&	1.172 $\pm$ 0.02	&	\textbf{1.120} $\pm$ \textbf{0.02}	&	1.159 $\pm$ 0.02	&	1.177 $\pm$ 0.02	&	\underline{\textbf{1.103}} $\pm$ \underline{\textbf{0.02}}	&	1.141 $\pm$ 0.02	\\
    \hline															

    sepal len. (2)	&	0.847 $\pm$ 0.02	&	0.848 $\pm$ 0.02	&	\textbf{0.816} $\pm$ \textbf{0.02}	&	0.832 $\pm$ 0.02	&	0.979 $\pm$ 0.02	&	\underline{\textbf{0.809}} $\pm$ 0.02	&	0.833 $\pm$ 0.02	\\
    sepal len. (3)	&	0.830 $\pm$ 0.02	&	0.847 $\pm$ 0.02	&	\underline{\textbf{0.816}} $\pm$ \underline{\textbf{0.02}}	&	0.826 $\pm$ 0.02	&	0.904 $\pm$ 0.02	&	\underline{\textbf{0.816}} $\pm$ \underline{\textbf{0.02}}	&	\textbf{0.821} $\pm$ \textbf{0.02}	\\
    sepal len. (4)	&	0.822 $\pm$ 0.02	&	0.840 $\pm$ 0.02	&	0.816 $\pm$ 0.02	&	0.828 $\pm$ 0.02	&	0.844 $\pm$ 0.02	&	\underline{\textbf{0.801}} $\pm$ \underline{\textbf{0.02}}	&	\textbf{0.813} $\pm$ \textbf{0.02}	\\
    \hline															

    sepal wid. (2)	&	0.450 $\pm$ 0.01	&	0.445 $\pm$ 0.01	&	\underline{\textbf{0.439}} $\pm$ \underline{\textbf{0.01}}	&	0.452 $\pm$ 0.01	&	0.523 $\pm$ 0.02	&	0.444 $\pm$ 0.01	&	\textbf{0.441} $\pm$ \textbf{0.01}	\\
    sepal wid. (3)	&	0.446 $\pm$ 0.01	&	0.451 $\pm$ 0.01	&	\underline{\textbf{0.439}} $\pm$ \underline{\textbf{0.01}}	&	\textbf{0.440} $\pm$ \textbf{0.01}	&	0.463 $\pm$ 0.01	&	\textbf{0.440} $\pm$ \textbf{0.01}	&	0.449 $\pm$ 0.01	\\
    sepal wid. (4)	&	0.445 $\pm$ 0.01	&	0.443 $\pm$ 0.01	&	\textbf{0.439} $\pm$ \textbf{0.01}	&	0.444 $\pm$ 0.01	&	0.442 $\pm$ 0.01	&	\underline{\textbf{0.437}} $\pm$ \underline{\textbf{0.01}}	&	0.445 $\pm$ 0.01	\\

\hline
\end{tabular}
\end{table}

\section{Parameter Selection Robustness}\label{section:paramselection}

\begin{figure}[!htb]
    \centering
    \includegraphics[width = 0.9\textwidth]{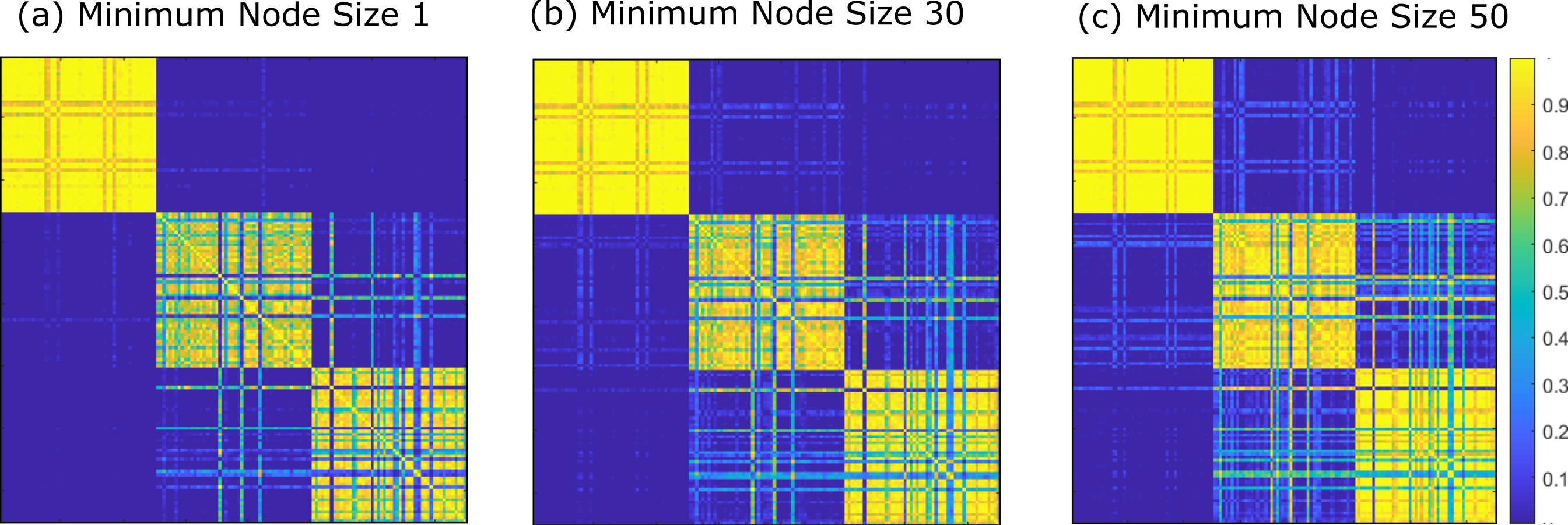}
    \caption{Scaled images of the proximity matrices of the Iris~\cite{anderson1936iris, UCI2019} data set setting the minimum node size (parameter \texttt{nodesize}) at different levels.  Increasing the minimum node size inflates the proximity values, as can be seen by viewing the increased proximities in (c) as compared to those in (a).}
    \label{fig:nodesize}
\end{figure}

In this section we show that RF-PHATE is robust to parameter tuning.  Random forests are robust to overfitting the data; increasing the number of trees can only improve random forest performance \cite{Cutler2012} and the random forest generalization error converges almost surely to a limit~\cite{rf}.  Random forests use, by default, fully-grown trees.  In the context of proximity generation, decreasing the number of terminal nodes (i.e. pruning the tree) or setting a minimum node size, inflates the proximity values since the probability of two observations falling into the same terminal node increases as the number of terminal nodes decreases. This is demonstrated in Figure \ref{fig:nodesize}.

It has been empirically shown~\cite{Cutler2012, uriarte2006} that the only  parameter to which random forests are sensitive is the number of randomly selected variables to be considered at a given terminal node (the parameter \texttt{mtry}). The \texttt{randomForest}~\cite{randomForest} default for classification is $\sqrt{n}$, where $n$ is the number of observations. For regression, the default is $n/3$. We selected values of \texttt{mtry} centered at the default to test RF-PHATE's robustness to this parameter.

\begin{figure}[!htb]
    \centering
    \includegraphics[width = 0.9\textwidth]{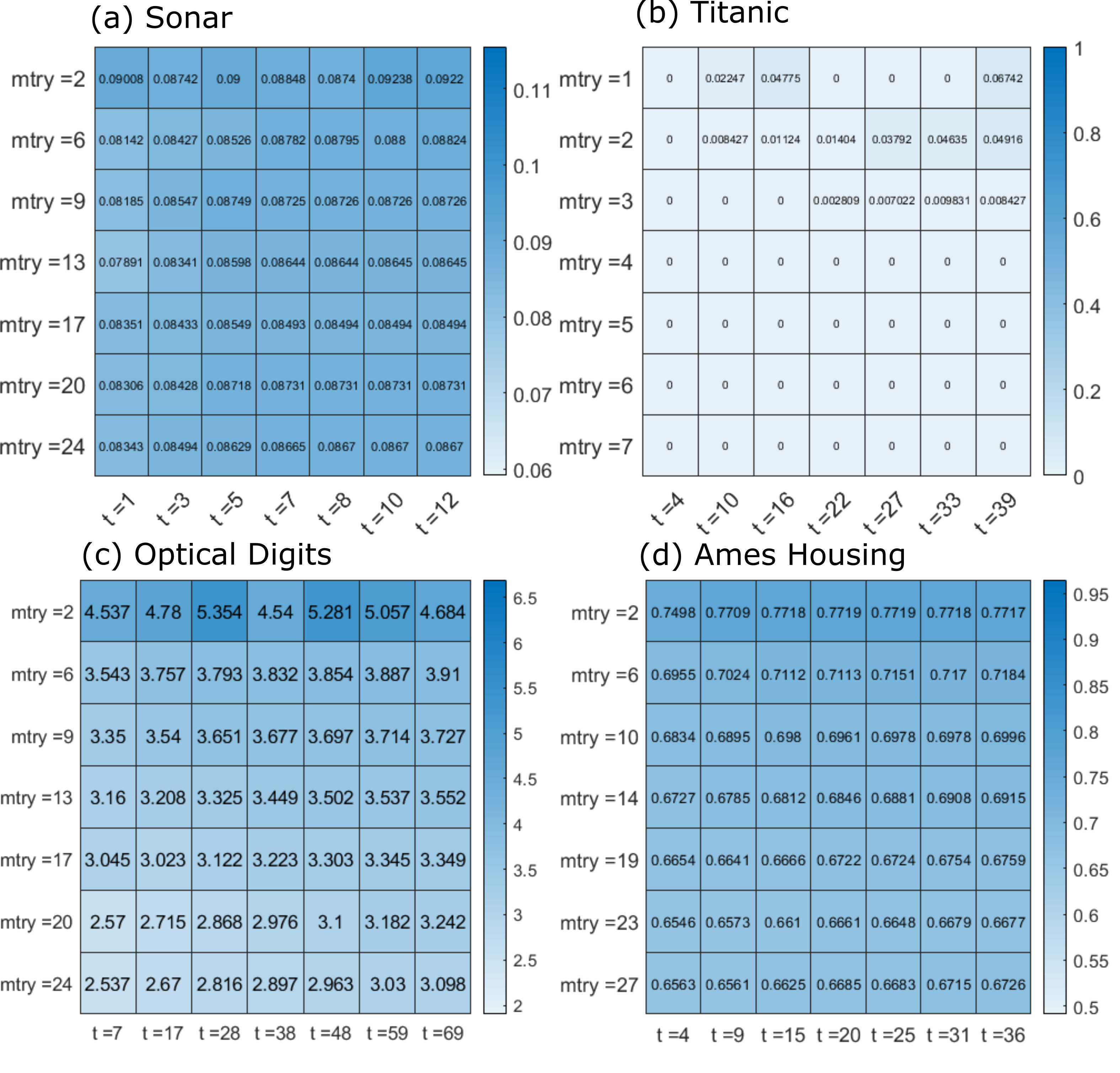}
    \caption{Heatmaps of the results of regression over the top important variable using the 2-dimensional RF-PHATE embeddings (the same metrics used in Tables 1 and 2) over a range of $t$ and \texttt{mtry} values centered at their defaults on the (a) \textit{Sonar}, (b) \textit{Titanic}, (c) \textit{Optical Digits}, and (d) \textit{Ames Housing} datasets. The results are  consistent across all values of of the parameters, demonstrating that RF-PHATE is robust to the choice of $t$ and \texttt{mtry}.}
    \label{fig:heatmaps}
\end{figure}

Diffusion-based dimensionality reduction is sensitive to the number of time steps, $t$, considered when conducting a ``random walk'' across all possible transitions~\cite{Moon2019, duque2019visualizing,gigante2019visualizing,horoi2020low}. We therefore test this parameter as well. The optimal $t$, as selected using VNE, serves as a center point to the values of $t$ considered for parameter tuning. The results are shown in Figure \ref{fig:heatmaps}. For each of the datasets in this figure, we ran RF-PHATE using ranges of \texttt{mtry} and $t$ centered about the default values.  We recorded the error rate (for categorical variables) or RMSE (for continuous variables) regressing on the top variable using the 2-dimensional RF-PHATE embeddings as features. This is the same metric used in Tables 1 and 2.

The results show that RF-PHATE is not sensitive to \texttt{mtry} or $t$ in regards to capturing important variables in a low-dimensional space with a categorical response. However, we have observed that high values of $t$ (much greater than the optimally selected $t$) tend to visually collapse clusters. We therefore recommend that the default choice of $t$, or a value close to it, should be used. 

\section{Embedding random forest proximities with other methods}
\label{sec:prox}

Because of its abilities to accurately preserve local and global structure, we chose to adapt the PHATE algorithm to embed the random forest proximities. Here we show the results when embedding the proximities using other  dimensionality reduction methods with varying success.  We compared results embedding the proximities using Isomap, LLE, and MDS in addition to our proposed method. These are each denoted as RF-``method''. 

\begin{figure}[!htb]
    \centering
    \includegraphics[width = 0.9\textwidth]{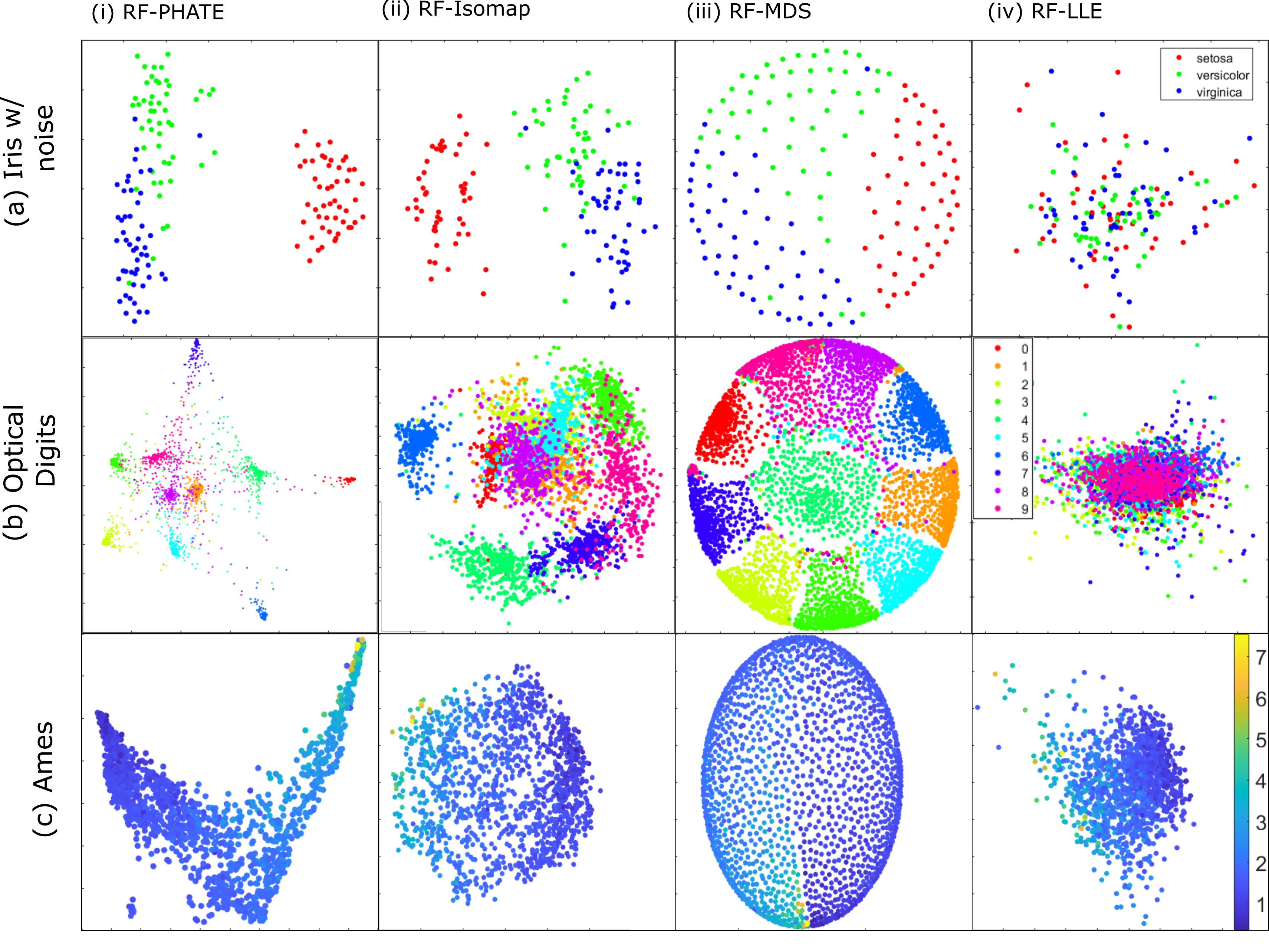}
    \caption{Visualizations using random forest proximities as a kernel for Isomap, MDS, and LLE on the (a) \textit{Iris} with noise,  (b) \textit{Optical Digits}, and (c) \textit{Ames Housing} data sets. For low-dimensional data sets, RF-Isomap (a)ii  gives meaningful results; however, it can be seen in (a)iii (b)iii and (c)iii that MDS applied to the proximities does not accurately capture the global structure of the data on  high dimensional data sets.  Isomap with proximities (a)ii and (b)ii captures some of the global structure but does not reduce noise. LLE with proximities does not produce meaningful results on (a)iv or (b)iv.}
    \label{fig:proxmethods}
\end{figure}

RF-Isomap gives meaningful results in small, low-dimensional data sets even in the presence of noise variables, but less desirable results for higher-dimensional data sets (see Figure \ref{fig:proxmethods}), while RF-LLE does not perform well without significant parameter tuning. For higher-dimensional data sets, including when noise variables are present, RF-MDS does not capture the geometric (global) structure of the data.  RF-Isomap captures more of the global structure, but does not sufficiently denoise the data. In contrast, RF-PHATE shows the global and local structure of the data, even when many of the variables are irrelevant for the supervised task.

\section{Computational Environment}\label{section:environment}

All computations were done using an Intel Xeon CPU E5-1650 v2 @ 3.50 GHz with 6 cores.  The random forest proximities were computing using  \texttt{R} version 3.6.2.  All other computations were done in Matlab 2019b.

\end{document}